\documentclass[11pt,round]{article}
\usepackage[preprint]{neurips_2022}
\usepackage[utf8]{inputenc} 
\usepackage[T1]{fontenc}    
\usepackage{amsfonts, amsmath}       
\usepackage{float}
\usepackage{graphicx} 
\usepackage{bold-extra} 
\usepackage{lmodern}
\usepackage[footnotesize]{caption} 
\usepackage{hyperref}
 \hypersetup{
     colorlinks=true,
     linkcolor=blue,
     filecolor=blue,
     citecolor = blue,      
     urlcolor=cyan,
     }
     
\usepackage{xpatch}
 
\usepackage{listings} 
\usepackage{tcolorbox}
\usepackage{realboxes}
\usepackage[export]{adjustbox}
\definecolor{mygreen}{rgb}{0,0.6,0}
\definecolor{mygray}{rgb}{0.5,0.5,0.5}
\definecolor{mymauve}{rgb}{0.58,0,0.82}
\definecolor{mygray}{rgb}{0.8,0.8,0.8}
\usepackage{placeins}

\makeatletter
\xpretocmd\lstinline{\Colorbox{mygray!40}\bgroup\appto\lst@DeInit{\egroup}}{}{}
\makeatother

\title{Interpreting Neural Networks through the Polytope Lens}
\author{Sid Black\thanks{equal contribution} \And Lee Sharkey$^\ast$\And Leo Grinsztajn\And Eric Winsor\And Dan Braun\And Jacob Merizian\And Kip Parker\And Carlos Ramón Guevara\And Beren Millidge\And Gabriel Alfour\And Connor Leahy}

\date{August 2022}

\begin{document}
\maketitle
\setcounter{footnote}{0} 

\begin{abstract}
    Mechanistic interpretability aims to explain what a neural network has learned at a nuts-and-bolts level. What are the fundamental primitives of neural network representations? What basic objects should we use to describe the operation of neural networks mechanistically? Previous mechanistic descriptions have used individual neurons or their linear combinations to understand the representations a network has learned. But there are clues that neurons and their linear combinations are not the correct fundamental units of description---directions cannot describe how neural networks use nonlinearities to structure their representations. Moreover, many instances of individual neurons and their combinations are polysemantic (i.e. they have multiple unrelated meanings). Polysemanticity makes interpreting the network in terms of neurons or directions challenging since we can no longer assign a specific feature to a neural unit.  In order to find a basic unit of description that doesn’t suffer from these problems, we zoom in beyond just directions to study the way that piecewise linear activation functions (such as ReLU) partition the activation space into numerous discrete polytopes. We call this perspective the ‘polytope lens’. Although this view introduces new challenges, we think they are surmountable and that more careful consideration of the impact of nonlinearities is necessary in order to build better high-level abstractions for a mechanistic understanding of neural networks. The polytope lens makes concrete predictions about the behavior of neural networks, which we evaluate through experiments on both convolutional image classifiers and language models. Specifically, we show that polytopes can be used to identify monosemantic regions of activation space (while directions are not in general monosemantic) and that the density of polytope boundaries reflect semantic boundaries. We also outline a vision for what mechanistic interpretability might look like through the polytope lens. 
\end{abstract}

\clearpage
{\small
\setcounter{tocdepth}{3}
\tableofcontents
}

\section{Introduction}

How should we carve a neural network at the joints? Traditionally, mechanistic descriptions of neural circuits have been posed in terms of neurons, or linear combinations of neurons also known as  ‘directions’. Describing networks in terms of these neurons and directions has let us understand a surprising amount about what they’ve learned \citep{cammarata2020}. But these descriptions often possess undesirable properties - such as polysemanticity and inability to account for nonlinearity - which suggest to us that they don’t always carve a network at its joints.

If not neurons or directions, then what should be the fundamental unit of a mechanistic description of what a neural network has learned? Ideally, we would want a description in terms of some object that throws away unnecessary details about the internal structure of a neural network while simultaneously retaining what’s important. In other words, we’d like a less \href{https://en.wikipedia.org/wiki/Leaky_abstraction}{\emph{‘leaky’ abstraction}} for describing a neural network’s mechanisms.

We propose that a particular kind of mathematical object – a ‘polytope’ – might serve us well in mechanistic descriptions of neural networks with piecewise-linear activations.\footnote{And, with some relaxations to ‘soft’ polytopes, the polytope lens might also let us mechanistically describe neural networks with activations such as GELU and Swish. Some prior work exists that extends the polytope lens to such activations \citep{balestriero2018}. See the Appendix C for further discussion.} We believe they might let us build less leaky abstractions than individual neurons and directions alone, while still permitting mechanistic understandings of neural networks of comparable length and complexity. 

To help explain how the polytope lens could underlie mechanistic descriptions of neural networks, we first look at the problems that arise when using individual neurons (both biological and artificial) and then when using directions as the basic units of description and suggest how this perspective offers a potential solution. 

\subsection{Are individual neurons the fundamental unit of neural networks?}

Studying the function of single neurons has a long history. The dominant view in neuroscience for approximately one hundred years was the ‘neuron doctrine’ \citep{yuste2015}. The neuron doctrine contended that the way to understand neural networks is to understand the responses  of individual neurons and their role in larger neural circuits. This led to significant successes in the study of biological neural circuits, most famously in the visual system. Early and important discoveries within this paradigm included cells in the frog retina that detect small patches of motion (fly detectors) \citep{lettvin1959}; cells in the visual cortex with small receptive fields that detect edges \citep{hubel1962}, cells in the higher visual system that detect objects as complex as faces \citep{sergent1992}, and many even highly abstract multimodal concepts appear to be represented in single neurons \citep{quiroga2005,quianquiroga2009}.

Given their historic usefulness in the study of \emph{biological} neural networks, individual neurons are a natural first place to start when interpreting \emph{artificial} neural networks. Such an approach has led to significant progress. Many studies have suggested that it is possible to identify single neurons that responded to single features \citep{szegedy2014,zhou2015,Bau2017,olah2017feature}. Analysis of small neural circuits has also been done by inspecting individual neurons \citep{cammarata2020curve,goh2021multimodal}.

Mathematically, it’s not immediately obvious why individual neurons would learn to represent individual features given that, at least in linear networks, the weights and activations can be represented in any desired basis. One suggestion for why this would happen is the ‘privileged basis’ hypothesis \citep{elhage2021mathematical, elhage2022superposition}. This hypothesis states that element-wise nonlinear activation functions encourage functionally independent input features to align with individual neurons rather than directions.

Despite both historical success and the privileged basis hypothesis, it turns out that in many circumstances networks learn features that don't perfectly align with individual neurons. Instead, there have been some suggestions that networks learn to align their represented features with directions \citep{olah2018the,saxena2019}. 

\subsection{Are directions the fundamental unit of neural networks?}

One of the main reasons to prefer directions over individual neurons as the functional unit of neural networks is that neurons often appear to respond to multiple, seemingly unrelated things. This phenomenon is called polysemanticity.\footnote{In neuroscience, the polysemanticity is called mixed selectivity \citep{fusi2016}} \cite{nguyen2016} (supplement) and \cite{olah2017feature} were perhaps the first to explicitly identify neurons that represent multiple unrelated features in convolutional image classifiers. Polysemantic neurons have also been found in large language models \citep{geva2021} and multimodal networks \citep{goh2021multimodal}, and in the brain \citep{tanabe2013}. They are usually found by looking at the dataset examples that maximally activate specific neurons and noticing that there are multiple distinct groups of features represented in the examples. Below are a few examples of polysemantic neurons from a convolutional image classifier (InceptionV1\footnote{We chose InceptionV1 since it has served as a kind of ‘model system’ in previous mechanistic interpretability work. But the Pytorch \href{https://github.com/pytorch/vision/blob/main/torchvision/models/googlenet.py}{implementation} of the InceptionV1 architecture (also known as GoogLeNet), it transpires, differs from the original. The original had no batch norm, whereas the Pytorch version does.}) and a large language model, GPT2-Medium.

\begin{figure}[!ht]
    \centering
    \includegraphics[width=0.55\linewidth]{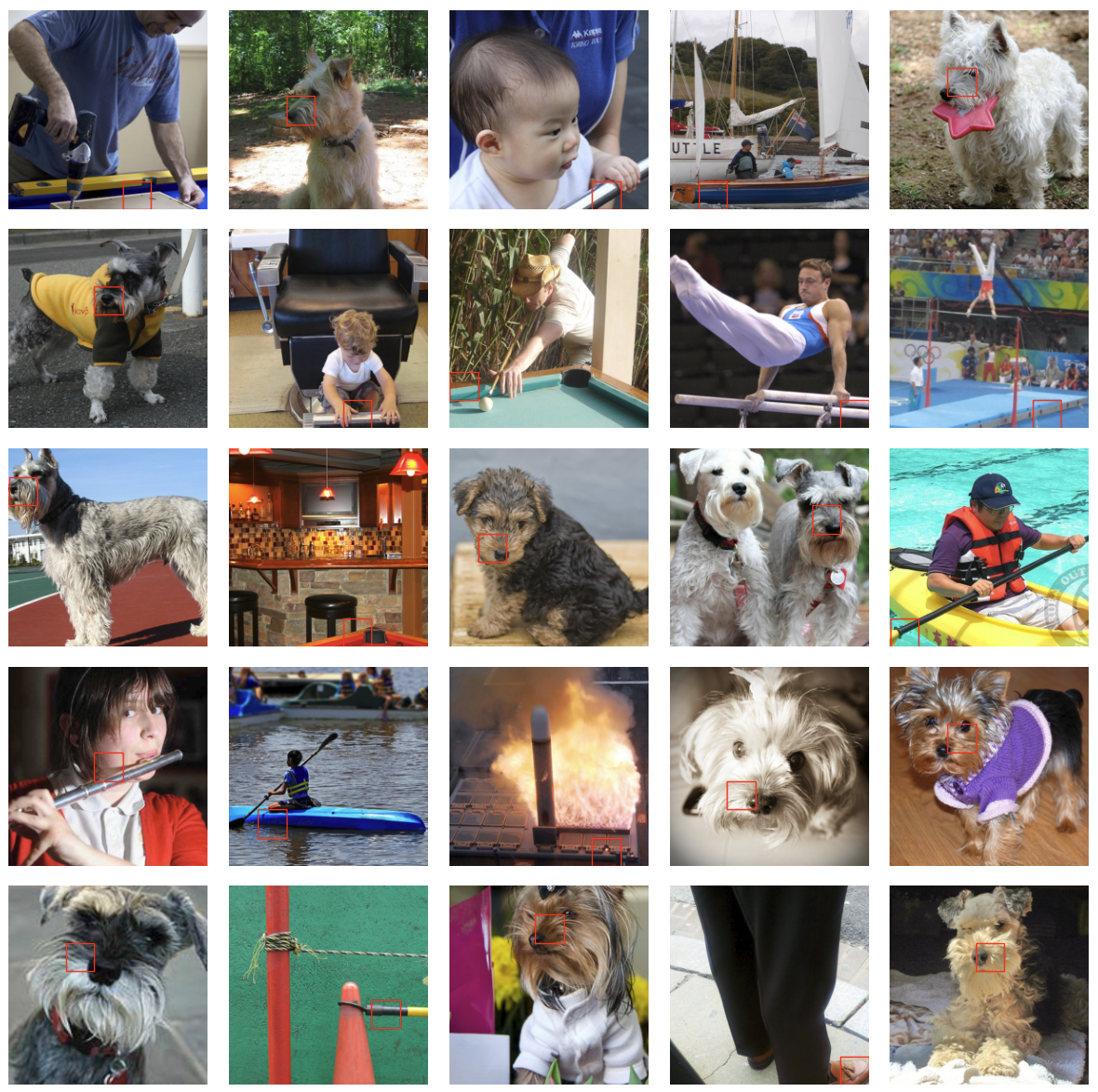}
    \caption{An example of a polysemantic neuron in InceptionV1 (layer inception5a, neuron 233) which seems to respond to a mix of dog noses and metal poles (and maybe boats).}
    \label{fig:Fig1}
\end{figure}

\begin{figure}[!ht]
    \centering
    \includegraphics[width=0.75\linewidth]{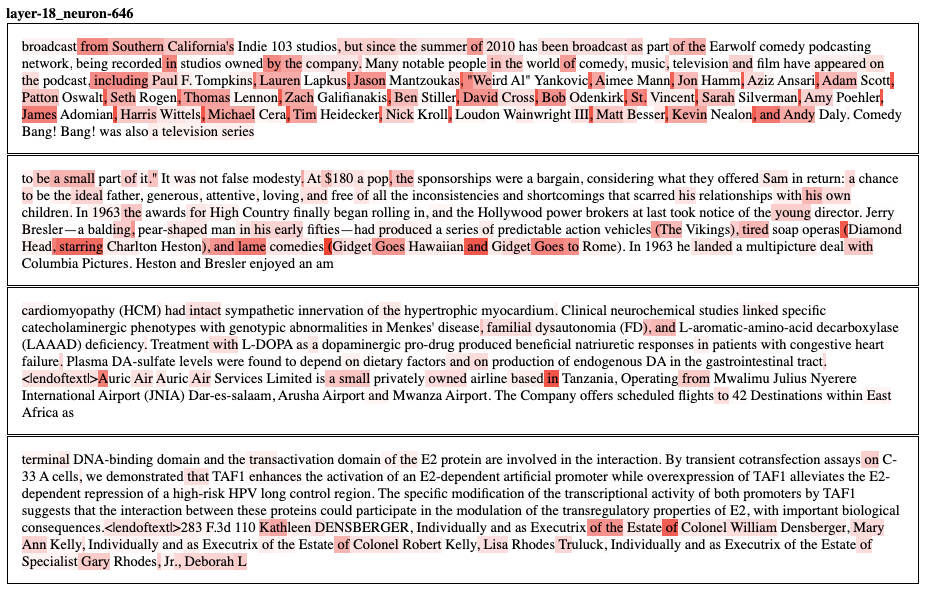}
    \caption{An example of a polysemantic neuron in GPT2-Medium. The text highlights represent the activation magnitude - the redder the text, the larger the activation. We can see that this neuron seems to react strongly to commas in lists, but also to diminutive adjectives (‘small’, ‘lame’, ‘tired’) and some prepositions (‘of’, ‘in’, ‘by’), among other features.}
    \label{fig:Fig2}
\end{figure}

One explanation for polysemantic neurons is that networks spread the representation of features out over multiple neurons. By using dimensionality reduction methods, it’s often possible to find directions (linear combinations of neurons) that encode single features, adding credence to the idea that directions are the functional unit of neural networks \citep{olah2018the,saxena2019a,mante2013}. This chimes with the ‘features-as-directions perspective’ \citep{elhage2022superposition}. Under this perspective, the magnitude of neural activations loosely encodes ‘intensity’ or ‘uncertainty’ or ‘strength of representation’, whereas the direction encodes the semantic aspects of the representation.\footnote{Similar encoding methods have been widely observed in neuroscience where they are called “population coding”. Population codes have been found or hypothesized to exist in many neural regions and especially the \href{https://www.ini.uzh.ch/~kiper/georgopoulos.pdf}{motor cortex}.} 

If there are fewer features than neurons (or an equal number of both), then each feature can be encoded by one orthogonal direction. To decode, we could simply determine which linear combination of neurons encodes each feature. However, if there are more features than neurons, then features must be encoded in non-orthogonal directions and can interfere with (or \href{https://en.wikipedia.org/wiki/Aliasing}{alias} - see Appendix D) one another. In this case, the features are sometimes said to be represented in ‘superposition’ \citep{elhage2022superposition}.\footnote{The idea that non-orthogonal representations interfere with each other has a long history in machine learning, starting with the study of the memory capacity of associative memories such as Hopfield networks which face the same underlying tradeoff between information capacity and orthogonality \citep{hopfield1982,abu-mostafa1985}. When features are encoded in non-orthogonal directions, the activation of one feature coactivates all feature directions sharing a non-zero dot product with it, leading to interference.} In superposition, networks encode more features than they have orthogonal basis vectors. This introduces a problem for a naive version of the features-as-directions hypothesis: Necessarily, some feature directions will be polysemantic! If we assume that representations are purely linear, then it’s hard to see how networks could represent features in non-orthogonal directions without interference degrading their performance. Neural networks use nonlinearities to handle this issue. \cite{elhage2022superposition} argue that a Rectified Linear Unit (ReLU) activation does this through thresholding: If the interference terms are small enough not to exceed the activation threshold, then interference is ‘silenced’! For example, suppose neuron A is polysemantic and represents a cat ear, a car wheel, and a clock face, and neuron B represents a dog nose, a dumbbell, and a car wheel. When neuron A and B activate together, they can cause a downstream car neuron to activate without activating neurons that represent any of their other meanings, so long as their pre-activations are below threshold. 

Beyond enabling polysemanticity, nonlinearities introduce a second problem for the features-as-directions viewpoint. The directions in each layer, caused by a direction in an earlier layer, are no longer invariant to scaling, as we would expect in a fully linear network. If we scale the activations in a particular layer in a fully linear network by some scalar multiple, we expect the class prediction to remain the same - as this is equivalent to scaling the output logits. However, if we scale the activations in a particular layer in a \emph{non}-linear network, some neurons in later layers may ‘activate’ or ‘deactivate’. (i.e. their preactivation goes above or below threshold). In other words, scaling directions in one layer can change the direction (and hence the features represented) in later layers!

\begin{figure}[!ht]
    \centering
    \includegraphics[width=0.75\linewidth]{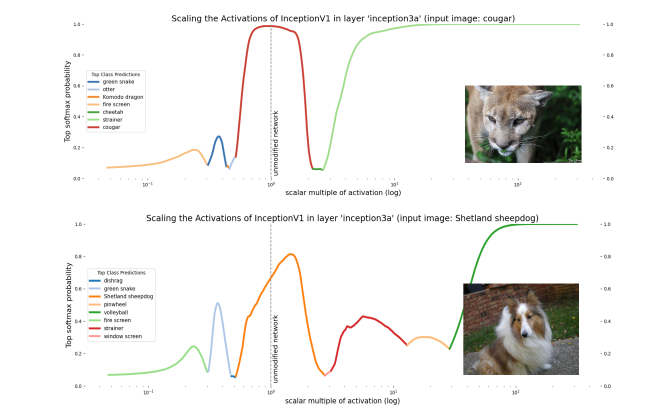}
    \caption{Scaling the activations in a layer causes semantic changes later in the network despite no change in activation direction in the scaled layer. The image on the right represents the input image.}
    \label{fig:Fig3}
\end{figure}

On the one hand, we should expect scaling the activation to change the direction in later layers. On the other, this poses a challenge to the features-as-directions view; scaling all representations relative to each other shouldn’t change their meaning except by changing their ‘intensity’. The naive version of the features-as-directions hypothesis requires the addition of something like a ‘distribution of validity’ within which directions represent the correct feature and outside of which they don’t. Unfortunately, the features-as-directions view doesn’t tell us what this distribution is. We’d like to know what the distribution is in order to know when our models might exhibit unpredictable out-of-distribution behavior. 

Despite these two limitations (polysemanticity and failure to be invariant to scale), the features-as-directions view has enabled much progress in understanding circuits of some neural networks, even permitting \cite{cammarata2021curve} to reverse engineer some circuits and reconstruct them by hand. So the view represents at least a substantial piece of the interpretability puzzle - and it seems true that \emph{some} directions carry a clear semantic meaning. Another reason to believe that the features-as-directions viewpoint is sensible is that, as we scale the hidden activations, neighbouring categories are quite often (but not always) semantically related. For instance, when we scale up the hidden layer activations for the cougar image, the network misclassifies it as a cheetah, which is still a big cat! 

Instead of radically overhauling the features-as-directions view, perhaps it only needs some modifications to account for the effects of nonlinearities, namely:
\begin{itemize}
    \item Invariances - We have shown that directions are not invariant to scaling. We want a modification that captures invariances in neural networks. For instance, we want something that points the way to ‘semantic invariances’ by identifying monosemantic components of neural networks even when subjected to certain geometric transformations (like scaling). 
    \item On/off-distribution - The features-as-directions view appears to be correct only when the scale of activations is within some permitted distribution. We want a way to talk about when activations are off-distribution with more clarity, which will hopefully let us identify regions of activation space where the behavior of our models becomes less predictable. 
\end{itemize}
To find an object that meets our needs, we turn to some recent developments in deep learning theory - a set of ideas that we call the ‘polytope lens’.

\section{The Polytope Lens}

Let’s consider an MLP-only network which uses piecewise linear activation functions, such as ReLU.\footnote{The arguments we make in support of the Polytope Lens  also apply to other activation functions such as GELU. But for simplicity we stick to piecewise linear activation functions because it’s easier to think geometrically in terms of straight lines rather than curvy ones.} In the first layer, each neuron partitions the input data space in two with a single hyperplane: On one side, the neuron is “on” (activated) and on the other side it’s “off”. 

On one side of the boundary, the input vector is multiplied by the weights for that neuron, which is just that neuron’s row of the weight matrix. On the other side, the input is instead projected to 0, as though that row of weight matrix were set to zero. We can therefore view the layer as implementing a different affine transformation on either side of the partition. For a mathematical description, see Appendix C. 

\begin{figure}[!ht]
    \centering
    \includegraphics[width=0.6\linewidth]{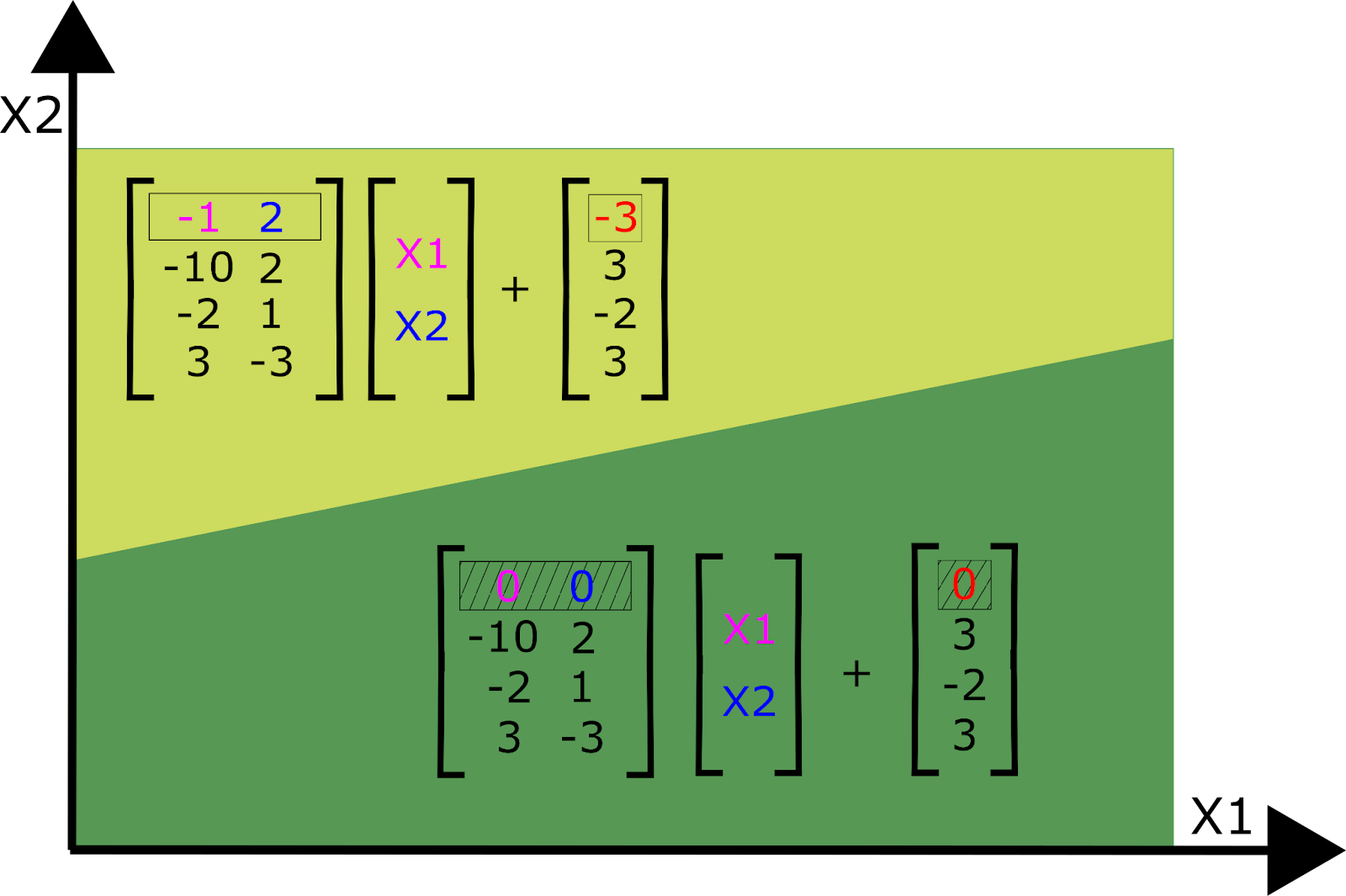}
    \caption{Affine transformations in the activated / unactivated regions of one neuron (assuming the three other neurons are activated).}
    \label{fig:Fig4}
\end{figure}

The orientation of the plane defining the partition is defined by the row of the weight matrix and the height of the plane is defined by the neuron’s bias term. The example we illustrate here is for a 2-dimensional input space, but of course neural networks typically have inputs that are much higher dimensional.

\begin{figure}[!ht]
    \centering
    \includegraphics[width=0.6\linewidth]{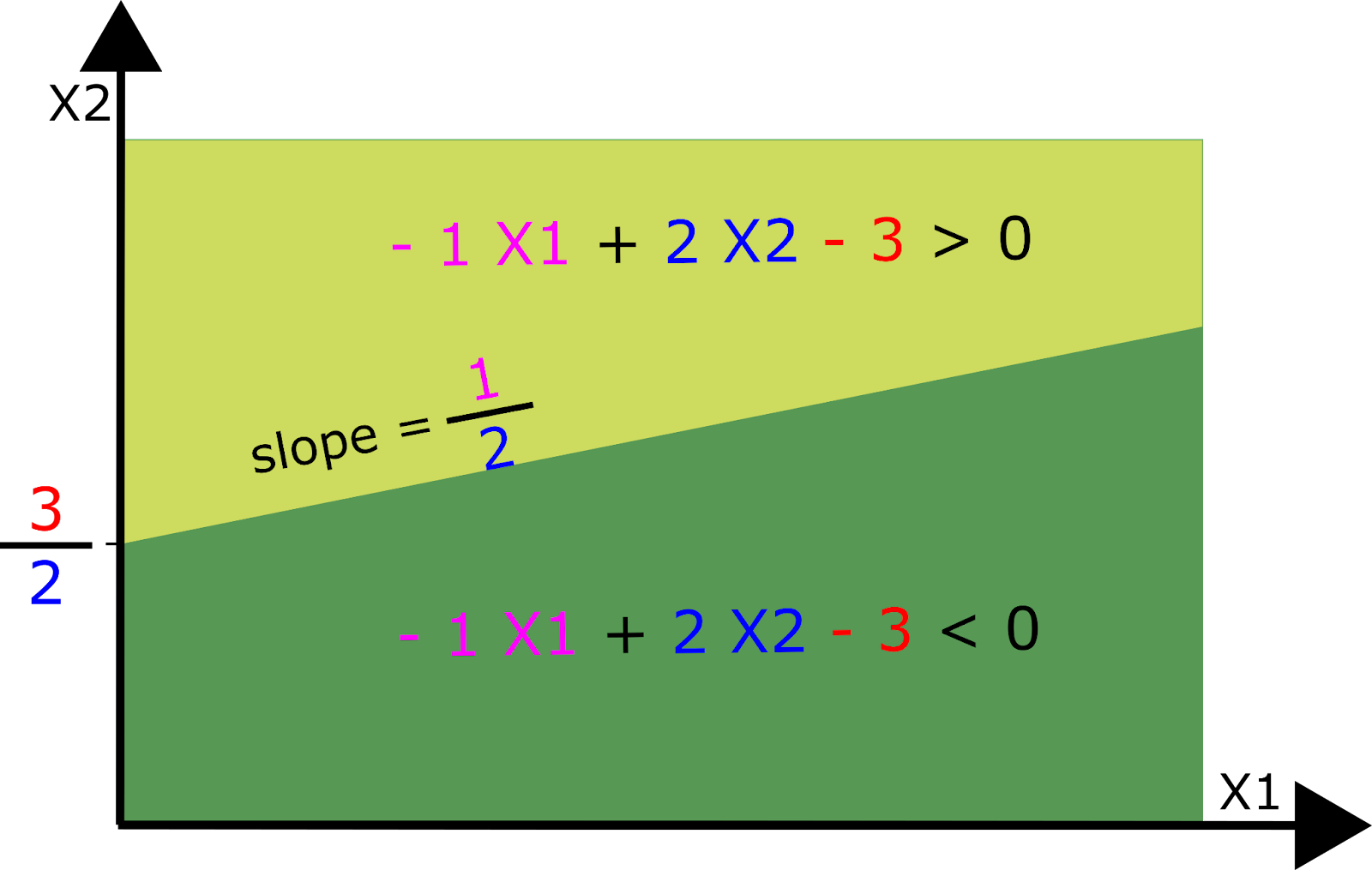}
    \caption{Polytope boundaries are defined by the weights and bias of a neuron. The weights determine the orientation of the (hyper-) plane and the bias determines its height.}
    \label{fig:Fig5}
\end{figure}

Considering all N neurons in layer 1 together, the input space is partitioned N times into a number of convex shapes called polytopes (which may be unbounded on some sides). Each polytope has a different affine transformation according to whether each neuron is above or below its activation threshold. This means we can entirely replace this layer by a set of affine transformations, one for each polytope. 

\begin{figure}[!ht]
    \centering
    \includegraphics[width=0.6\linewidth]{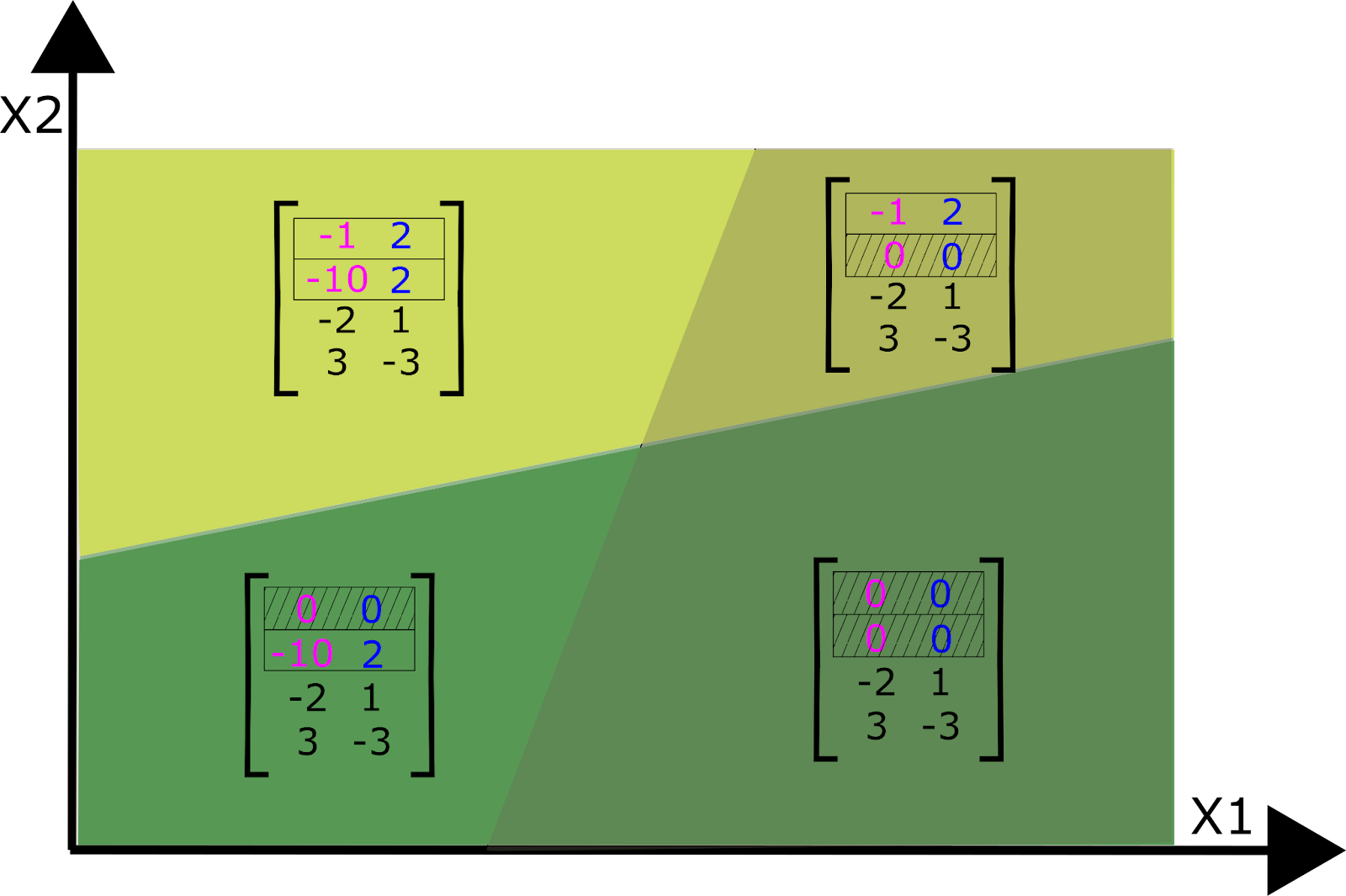}
    \caption{Four polytopes corresponding to four different affine transformations defined by two neurons in layer 1.}
    \label{fig:Fig6}
\end{figure}

As we add layers on top of layer 1, we add more neurons and, thus, more ways to partition the input space into polytopes, each with their own affine transformation. Thus, neural networks cut up the network’s input space into regions (polytopes) that each get transformed by a different set of affine transformations. Adding subsequent layers permits partition boundaries that \emph{bend} when they intersect with the partition boundaries of earlier layers \citep{hanin2019b}. The boundaries bend in different ways depending on the weights of the neurons in later layers that activate or deactivate. 

Each polytope can thus be analyzed as a fully linear subnetwork composed of a single affine transformation. Within each of these subnetworks, we would expect to see a set of interpretable directions that are scale invariant within each polytope. But the same directions in a different subnetwork might yield different interpretations. However, we should expect nearby polytope regions (subnetworks) to share similar affine transformations, and therefore similar semantics. We’ll discuss this further in the next section.

\begin{figure}[!ht]
    \centering
    \includegraphics[width=0.6\linewidth]{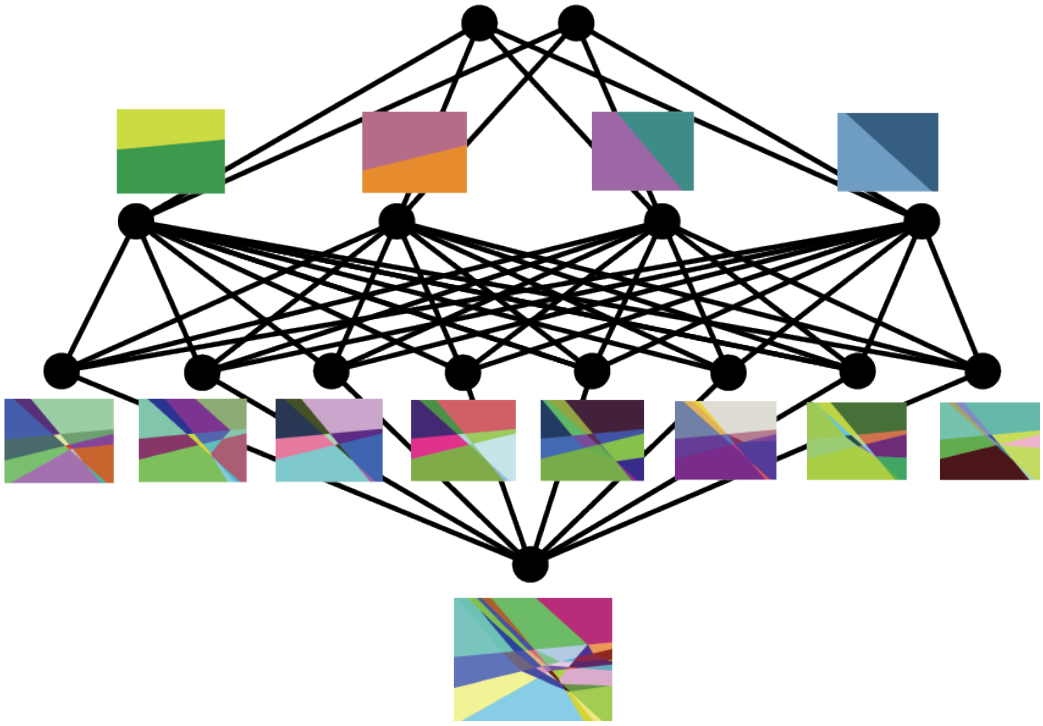}
    \caption{Image from \cite{hanin2019a}}
    \label{fig:Fig7}
\end{figure}

The polytope lens draws on some recent work in deep learning theory, which views neural networks as \textbf{max-affine spline operators} (MASOs) \citep{Balestriero2018b}. For a mathematical description of the above perspective, see Appendix C. 

The picture painted above is, of course, a simplified model of a far higher dimensional reality. When we add more neurons, we get a lot more hyperplanes and, correspondingly, a lot more polytopes! Here is a two dimensional slice of the polytopes in the 40768-dimensional input space of inception5a, with boundaries defined by all the subsequent layers:

\begin{figure}[!ht]
    \centering
    \includegraphics[width=0.75\linewidth]{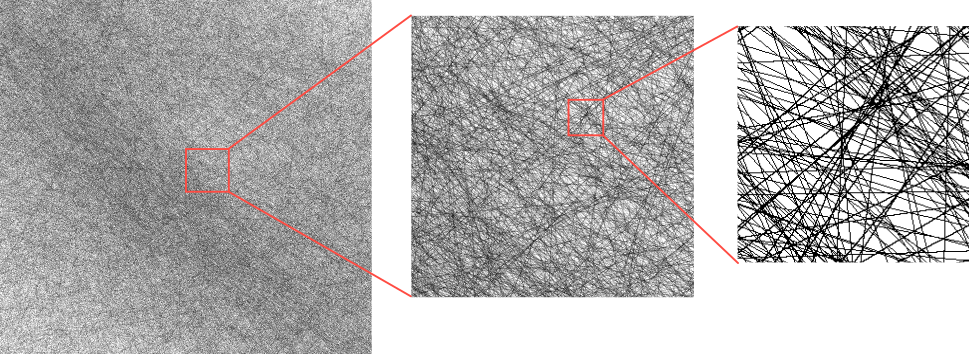}
    \caption{This figure depicts the polytope boundaries that intersect with a two-dimensional slice through the $832 * 7 * 7 = 40768$-dimensional input space of InceptionV1 layer inception5a. The slice was defined using the activation vectors caused by three images, one of a banana, a coffee cup, and a projector. The boundaries are defined using all neurons from inception5a to the classification logits. There are many polytopes in high dimensional space. If we instead used a lower layer, e.g. inception3a, then there would be many, many more polytope boundaries.}
    \label{fig:Fig8}
\end{figure}

In fact, as we add neurons, the number of polytopes the input space is partitioned into grows exponentially.\footnote{Although exponential, it’s not as many as one would naively expect - see \cite{hanin2019b}.} Such large numbers of polytopes become quite hard to talk about! Fortunately, each polytope can be given a unique code, which we call a ‘\textbf{spline code}’, defined in the following way: Consider the sequence of layers from $L$ to $L+K$. These layers define a set of polytope boundaries in the input space to layer $L$. A polytope’s spline code is simply a binary vector of length $M$ (where $M$ is the total number of neurons in layers $L$ to $L+K$) with a $1$ where the polytope causes a neuron to activate above threshold and $0$ otherwise. Notice that we can define a code for any sequence of layers; if we define a spline code from layer $L$ to $L+K$, the codes correspond to the polytopes that partition layer $L$’s input space. There is therefore a duality to spline codes: Not only are they a name for the region of input activation space contained within each polytope, but they can also be viewed as labels for pathways through layers $L$ to
$L+K$. 

\begin{figure}[!ht]
    \centering
    \includegraphics[width=0.75\linewidth]{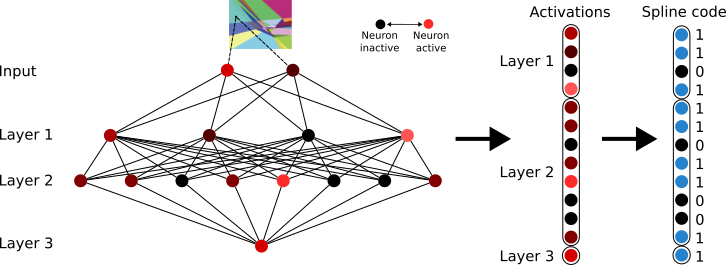}
    \caption{How spline codes are constructed in an MLP with ReLU activation functions. Activations in a set of layers are binarised according to whether each neuron is above or below threshold. (Partly adapted from \cite{hanin2019a}}
    \label{fig:Fig9}
\end{figure}

At least for deep ReLU networks, polytopes provide a mathematically correct description of how the input space is partitioned, unlike the naive version of the features-as-directions view which ignores the nonlinearities. However, polytopes are far more difficult to reason about than directions. They will need to give us greater predictive power to be worth the cost.

\subsection{Polytopes as the atoms of neural networks \& polytope regions as their molecules}

In the previous section, we discussed how it’s possible (in theory) to replace an entire ReLU network with each polytope’s affine transformation. Hence, polytopes provide a complete description of the input-output map of the network. Any inputs that belong to the same polytope are subject to the same affine transformation. In other words, the transformation implemented by the network is \emph{invariant within a polytope}.

But the invariance goes even further than individual polytopes; nearby polytopes implement similar transformations. To see why, consider two polytopes that share a boundary. Their spline codes differ by \emph{only one} neuron somewhere in the network turning on or off - in other words, the pathway taken by the activations through the network is identical except for the activation status of one neuron. Therefore, assuming the weights of some neurons aren’t unusually large, polytopes that have similar spline codes implement similar transformations in expectation.\footnote{This could be quantified, for instance, as the Frobenius norm of the difference matrix between the implied weight matrices of the affine transformations implemented in each polytope.} Hamming distance in the space of spline codes thus corresponds to expected distance in transformation space. 

It’s easy to see how this might be useful for semantics: If a network needs two similar-meaning inputs to be transformed similarly, all it needs to do is to project the inputs to nearby polytopes in hidden activation space. Here, the fundamental unit of semantics in the network, which we might call a feature, is a group of nearby polytopes that implement similar transformations. Notice that the addition of polytopes only modifies the features-as-directions view without replacing it entirely: Vectors in nearby polytopes usually share high cosine similarity, so ‘similar directions’ will correlate with ‘nearby polytopes’. Moreover, within a polytope the two views are identical.

This lets us make a few testable predictions about the relationship between semantics and polytope boundaries:

\begin{itemize}
    \item Prediction 1: \emph{Polysemantic directions overlap with multiple monosemantic polytope regions.}
    \begin{itemize}
        \item The polytope lens makes a prediction about how polysemanticity is implemented in neural networks: The multiple meanings of the polysemantic direction will correspond to monosemantic regions that have nonzero inner product with that direction.
    \end{itemize}
    \item Prediction 2: \emph{Polytope boundaries reflect semantic boundaries} 
    \begin{itemize}
        \item     Networks will learn to place more polytope boundaries between inputs of different classes than between the same classes. More generally, networks will learn to have regions denser with polytope boundaries between distinct features than between similar features.  
    \end{itemize}    
    \item Prediction 3: \emph{Polytopes define when feature-directions are on- and off-distribution.}
    \begin{itemize}
        \item Scaling hidden activation vectors eventually causes the prediction made by a classifier to change. It should be unsurprising that scaling the activations vectors of a nonlinear network well outside their typical distribution causes the semantics of directions to break. But neither the features-as-directions perspective nor the superposition hypothesis suggest what this distribution actually is. The polytope lens predicts that polytope boundaries define this distribution. Specifically, the class prediction made by the network should tend to change when the activation vector crosses a region of dense polytope boundaries. 
    \end{itemize}
\end{itemize}

We find that evidence supports predictions 1 and 2, and prediction 3 appears to be only partially supported by evidence. 

\subsubsection{Prediction 1: Polysemantic directions overlap with multiple monosemantic polytope regions}

Our approach to understanding polysemantic directions is to instead begin by identifying something in a network that \emph{is} monosemantic and work our way out from there, rather than starting with polysemantic directions and trying to figure out how they work. So, what \emph{is} monosemantic in a neural network? 

Neural networks implement approximately smooth functions, which means that small enough regions of activation space implement similar transformations. If similar representations are transformed in similar ways, it is likely that they “mean” similar things. This implies that small enough regions of activation space should be monosemantic, and indeed - this is why techniques like nearest-neighbor search work at all. To verify this claim, here we collect together activations in a) the channel dimension in InceptionV1 and b) various MLP layers in GPT2 and cluster them using HDBSCAN, a hierarchical clustering technique.\footnote{While we use HDBSCAN in this work, the specific algorithm isn't important. Any clustering algorithm that groups together any sufficiently nearby activations or codes should yield monosemantic clusters.} We observe that the majority of clusters found are monosemantic in both networks. For example, we observe clusters corresponding to specific types of animal in inception4c, and clusters responding to DNA strings, and specific emotional states in the later layers of GPT2-small. See Appendix E for more examples. 

\begin{figure}[!ht]
    \centering
    \includegraphics[width=0.75\linewidth]{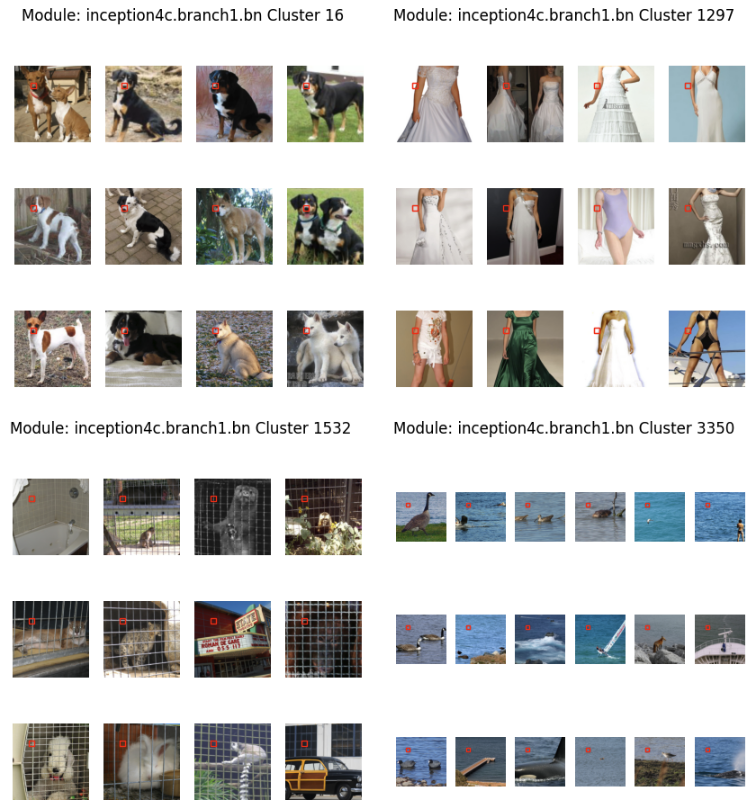}
    \caption{Examples of clusters of activations in the output of the first branch of the 4c layer of inceptionV1. For each cluster, we plot the images and hyperpixel corresponding to the activations. Clusters were computed with HDBSCAN on the activations for one spatial dimension, and randomly chosen among clusters containing enough images.}
    \label{fig:Fig10}
\end{figure}

\begin{figure}[!ht]
    \centering
    \includegraphics[width=0.39\linewidth]{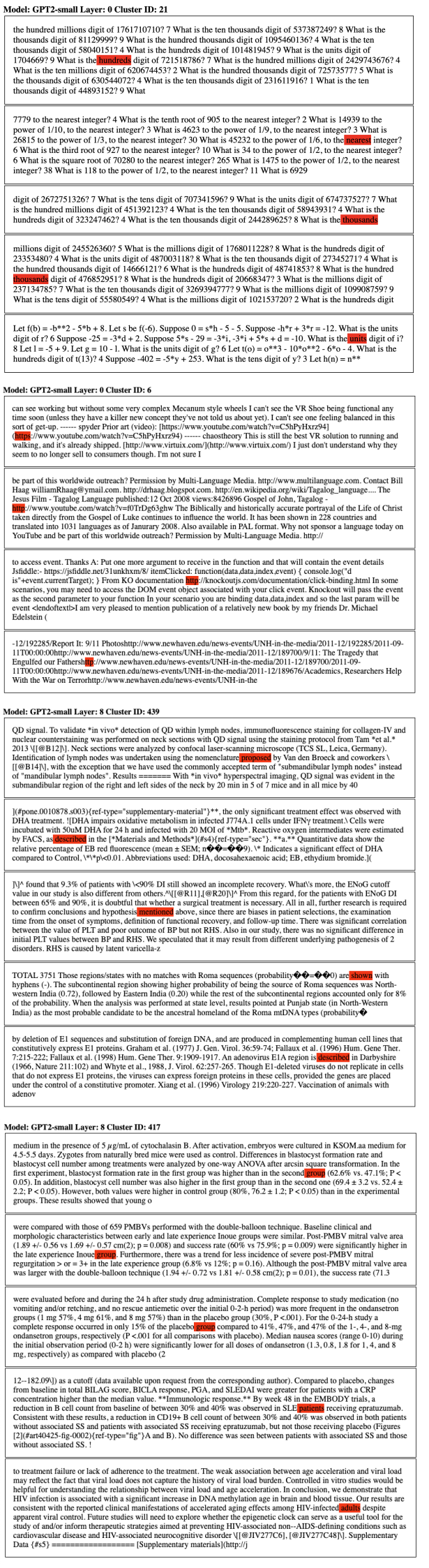}
    \caption{Dataset examples of clusters in the pre-activations of the MLP in various layers of GPT2-small. Clusters were computed using HDBSCAN on a random sample of the pile’s test set. Each token in the test set is treated as a separate point for clustering, and the specific token that has been clustered has been highlighted in red in each instance. We observe clusters responding both to specific tokens, and semantic concepts (typically, but not exclusively, in the later layers).}
    \label{fig:Fig11}
\end{figure}

Instead of finding monosemantic regions by clustering activations, it’s also possible to find them by clustering spline codes. This is mildly surprising, since we’ve ostensibly removed all information about absolute magnitude - and yet it’s still possible to group similar-meaning examples together. However, a single spline code implicitly defines a set of linear constraints. These constraints, in turn, describe a set of bounding hyperplanes which confine the set of possible activations to a small region in space. Thus, much of the information about the magnitude is still retained after binarization.

\begin{figure}[!ht]
    \centering
    \includegraphics[width=0.39\linewidth]{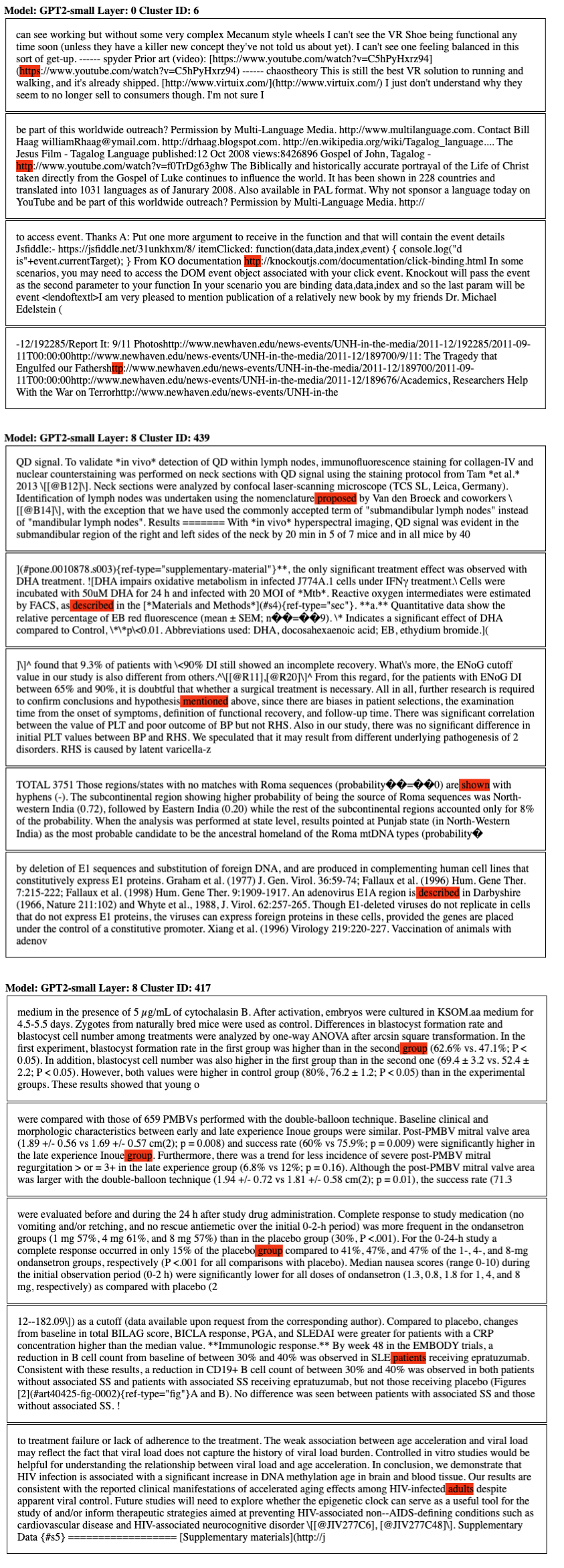}
    \caption{Dataset examples of clusters in the pre-activations of the MLP in various layers of GPT2-small. Clusters were computed using HDBSCAN on a random sample of the pile’s test set. The distance matrix for clustering in the above examples was computed using hamming distance on the binarized spline codes. Each token in the test set is treated as a separate point for clustering, and the specific token that has been clustered has been highlighted in red in each instance. We observe specific clusters in earlier layers that appear to be related to detokenization - i.e grouping “http” and “https” together. Clusters later layers tend to respond to higher level semantics - synonyms for groups of patients in medical trials, for example.}
    \label{fig:Fig12}
\end{figure}

We were interested in seeing if we would observe a similar effect with direction vectors found using dimensionality reduction techniques such as PCA or NMF. In theory, such directions should be those which explain the highest proportions of variance in the hidden space, and we would thus expect them to be amongst the most semantically consistent (monosemantic) ones.

In a “strong” version of the polytope lens - we might expect to see that even these directions, that we should expect to be monosemantic, also cross many polytope boundaries, potentially causing them to have different semantics at different magnitudes. However, the polytope lens does not preclude linear features - meaningful single directions are still possible in the latent space of a network with nonlinearities. To frame this in terms of paths through the network - it may be that there are linear features that are shared by all or most sets of paths.

To test this, we took the activations for a set of examples from a hidden layer (in this case, layer 4) of GPT2-small, and binarized them to get their spline codes. We then clustered the codes using HDBSCAN, with the same parameters as earlier experiments. Separately, we ran NMF on the raw activations (with 64 components) to find a set of directions. For each NMF vector, we measure the cosine similarity between it and each activation sample that we clustered, and plot the histograms in the below plots. The colours represent the cluster label that each activation has been assigned, each of which we have labelled with a semantic label by looking at the set of corresponding input samples. Since there are many clusters with extremely small cosine similarities that we are not interested in, we manually restrict the $x$-axis for each plot and display only the points with the largest similarities.

\begin{figure}[!ht]
    \centering
    \includegraphics[width=0.75\linewidth]{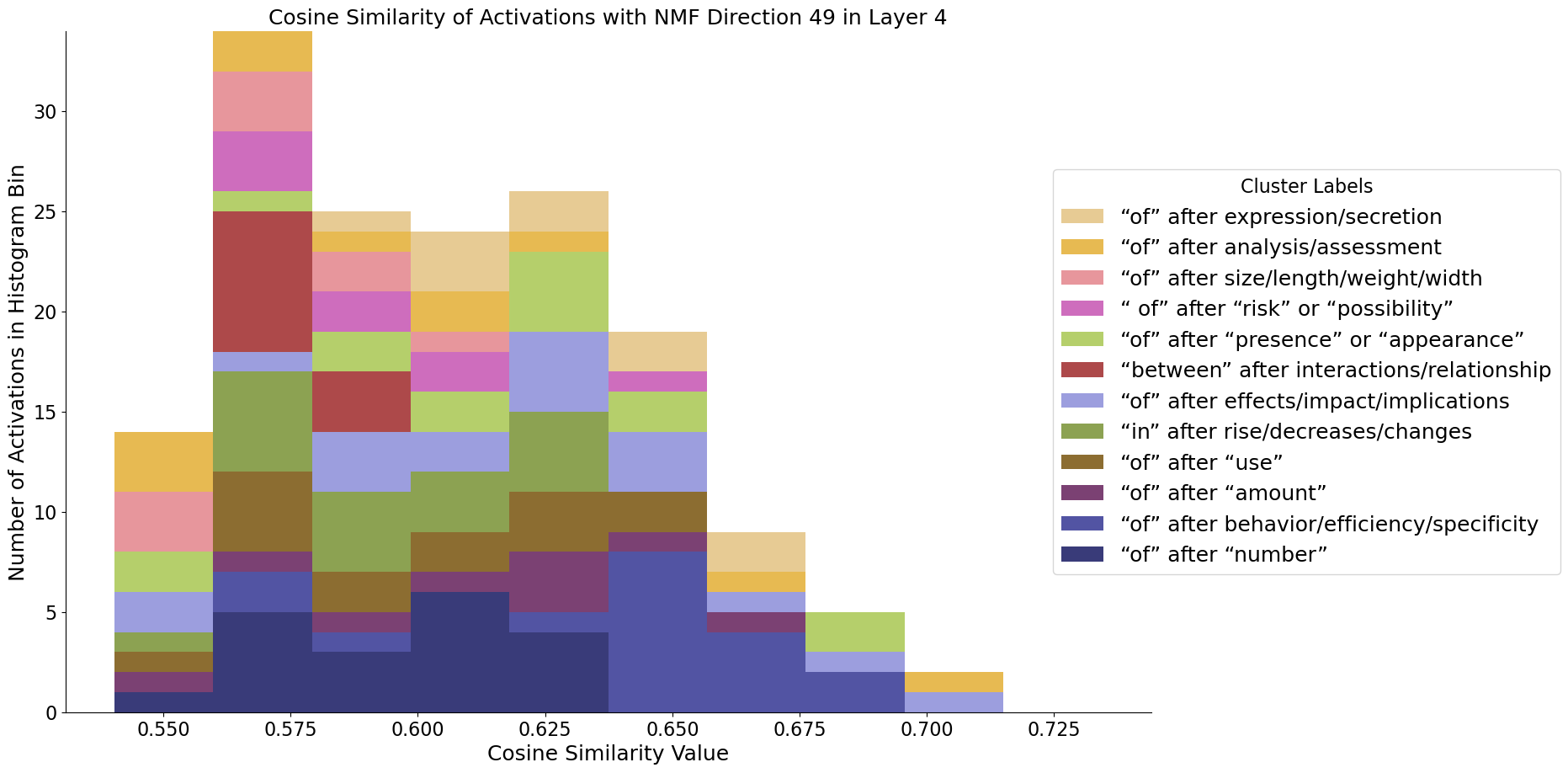}
    \caption{Cosine similarities with respect to NMF direction 49. Activations taken from the MLP in layer 4 of GPT2-Small, using data from The Pile's test set . The dataset examples with the highest cosine similarities are shown and coloured by their cluster label (ignoring the smallest clusters).}
    \label{fig:Fig13}
\end{figure}
It turns out that the directions found using NMF do appear to be largely monosemantic - so both models observed do seem to use features associated with directions to some extent, even if the basis directions still appear highly polysemantic. Using the same procedure, we can also find these monosemantic directions in InceptionV1: 

\begin{figure}[!ht]
    \centering
    \includegraphics[width=0.5\linewidth]{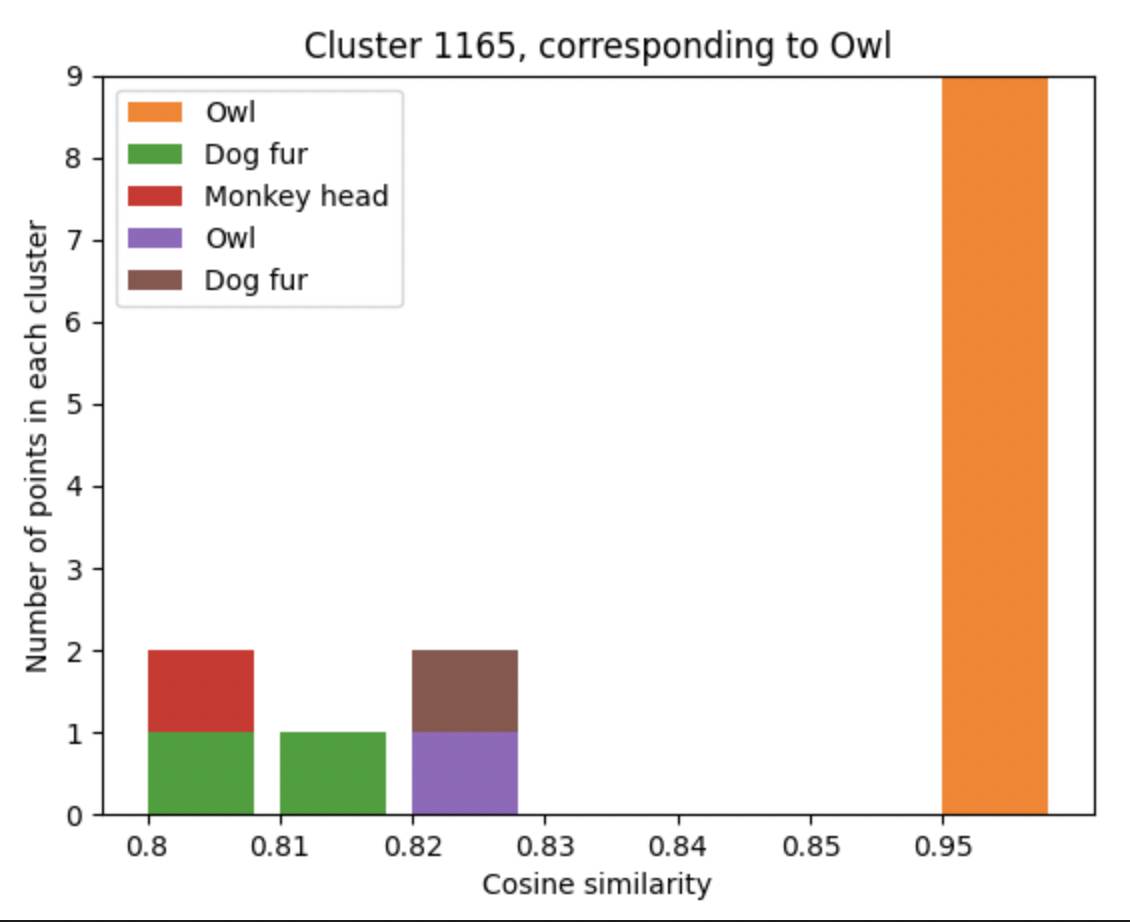}
    \caption{}
    \label{fig:Fig14}
\end{figure}
\begin{figure}[!ht]
    \centering
    \includegraphics[width=0.5\linewidth]{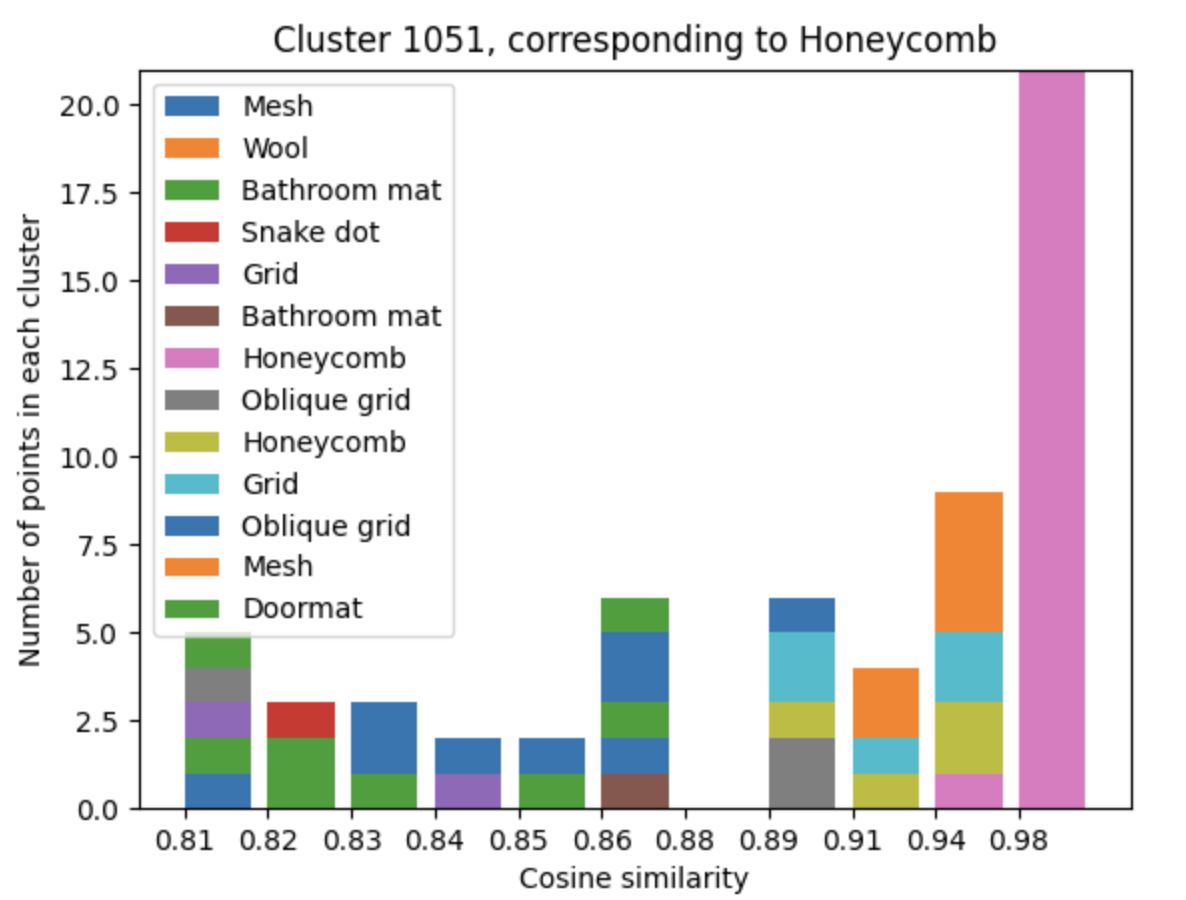}
    \caption{}
    \label{fig:Fig15}
\end{figure}

The above experiments suggest that there do exist feature directions which are coherent across all polytopes in some specific layer - meaning that the affine transformations formed across the set of all polytopes are sufficiently similar to some extent.

\subsubsection{Prediction 2: Polytope boundaries reflect semantic boundaries}

Why should we expect polytope boundaries to reflect semantic boundaries? One geometric intuition underlying this idea is that nonlinearities are needed to silence interference between non-orthogonal features in superposition. Polytope boundaries should therefore be placed between non-orthogonal feature directions so that activations in one feature direction don’t activate the other when they shouldn’t. Another intuition is that neural networks are often used in situations where outputs are not linearly separable functions of the inputs, such as image classification. To solve such tasks, neural networks fold and squeeze the input data manifold into a shape that is linearly separable in subsequent layers \citep{keup2022}. Affine transformations on their own cannot improve linear separability - but since a ReLU activation maps negative values to zero, it can be thought of as making a fold in the data distribution, with the position of the dent being controlled by the previous transformation’s weights. Several ReLU neurons in combination can also act to expose inner class boundaries - making classification in later layers possible where it wasn’t in earlier ones - by “folding” regions of the distribution into new, unoccupied dimensions (see the figure below for a 1D geometric interpretation). For this reason we may expect to see a concentration of ReLU hyperplanes around such distributions, as the network acts to encode features for later layers.

\begin{figure}[!ht]
    \centering
    \includegraphics[width=0.75\linewidth]{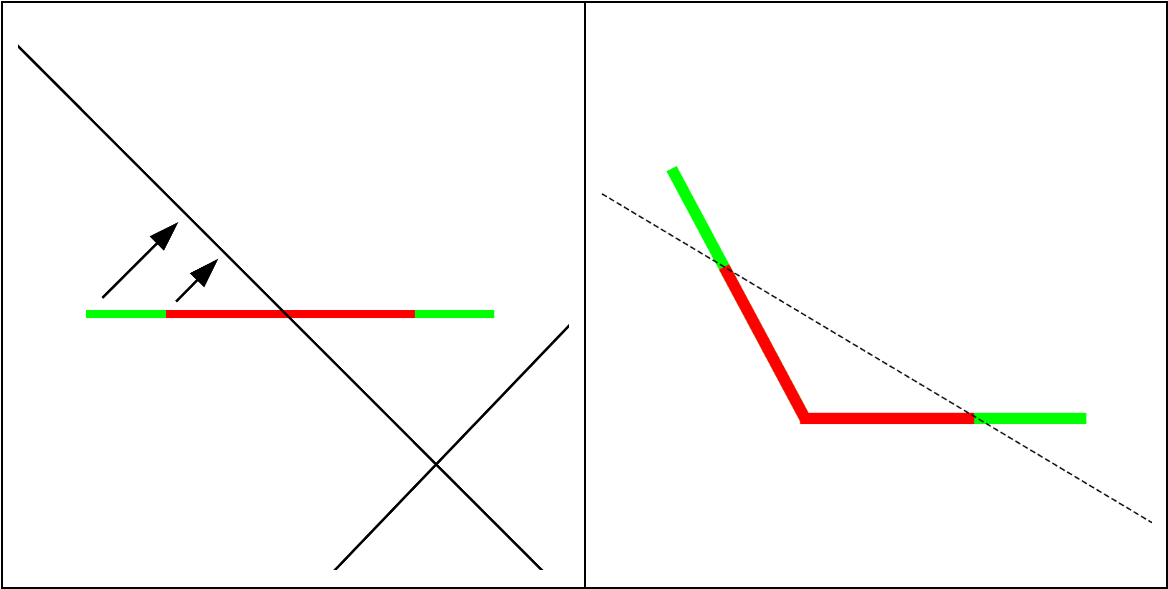}
    \caption{How a ReLU neuron can act to expose a previously non-separable class boundary, reproduced from Keup and Helias, (2022). The solid black line is a ReLU hyperplane, and the dashed line represents a potential decision boundary. In higher dimensions - this effect requires several ReLU hyperplanes to act in conjunction.}
    \label{fig:Fig16}
\end{figure}

Images of different classes will have many different features. Therefore, according to the polytope lens, activations caused by images from different classes should be separated by regions of denser polytope boundaries than those caused by images from the same class. Can we see this by looking at heat map visualizations of polytope density? Unfortunately, the network has too many neurons (and thus too many boundaries) to observe any differences directly.

\begin{figure}[!ht]
    \centering
    \includegraphics[width=0.65\linewidth]{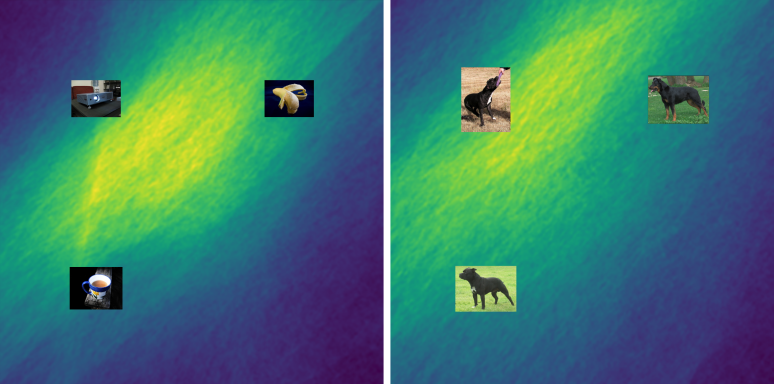}
    \caption{Heat maps of polytope density in a 2-D slice through the 40,768-dimensional input space of layer inception5a. The 2-D slice was made such that the activation vectors of three images lie on the plane. Then we calculated the spline codes (using layers inception5a to the output) for every point in a 4,096$\,\times\,$4,096 grid. Then we computed the Hamming distance between codes in adjacent pixels and applied a Gaussian smoothing. Observe that the densest region is the part of the image separating the three inputs. Method adapted from \cite{novak2018}, who calculated similar images for small networks.}
    \label{fig:Fig17}
\end{figure}

But when we measure the polytope densities directly (by dividing the distance between two activation vector’s spline codes by their Euclidean distance, it indeed turns out to be the case that activations caused by images of different classes are separated by regions denser in polytope boundaries than activations caused by images of the same class:

\begin{figure}[!ht]
    \centering
    \includegraphics[width=0.75\linewidth]{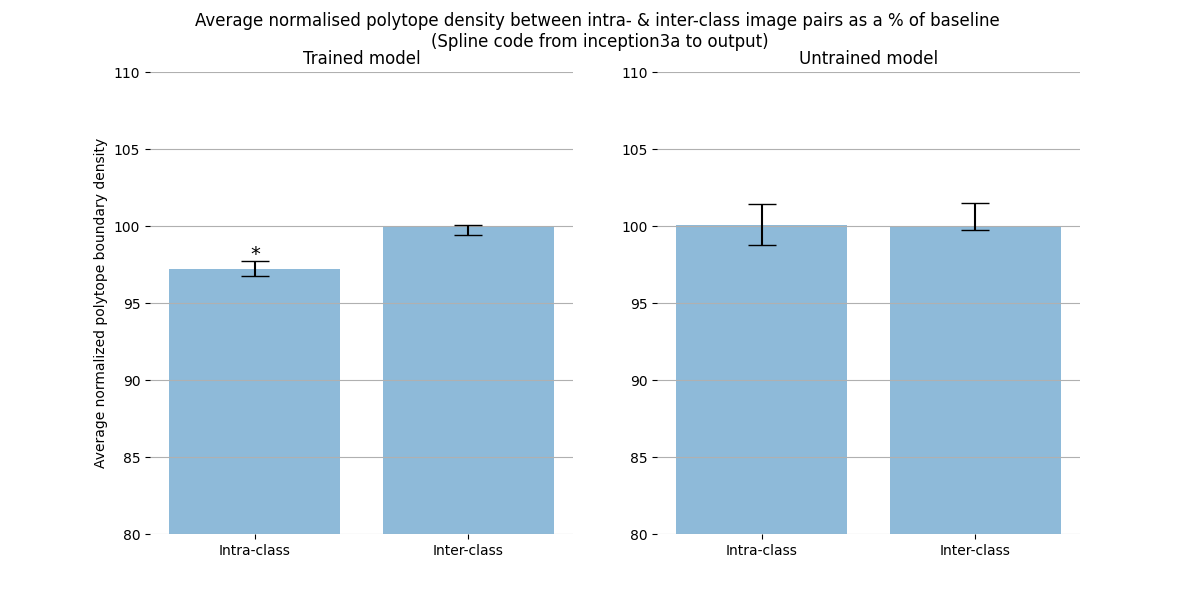}
    \caption{The average normalized polytope boundary density between the activation vector caused by images of the same or different classes. The left plot is for a trained network; the right an untrained network. Since images of different classes will also produce distant activations, we should consider the density of polytope boundaries rather than the absolute number of polytope boundaries between the activations produced by different images. To calculate the polytope boundary density between two points, we simply divide the Hamming distance in between their spline codes by the Euclidean distance between them. The polytope densities are normalized by dividing by the average polytope density between all pairs of vectors (both intra and inter class). Only for the trained network is the intra-class polytope density lower. This difference increases as we move higher in the network (Figure below). The error bars are 99\% bootstrapped confidence intervals. A single asterisk indicates a statistically significant difference according to a Welch's $t$-test ($t(1873.3)=-14.7$; $p=3.8e-46$). Note the $y$-axis begins at 80\%; the difference is small, but significant. We see a similar story when we interpolate (instead of simply measuring the total distances) between two images of the same or different classes (Appendix A).}
    \label{fig:Fig18}
\end{figure}

The intra- and inter-class difference is small, but significant. The difference gets more robust as we look at higher layers. The polytope lens predicts this because activations in lower layers represent low level features, which are less informative about image class than features in higher layers. For example, two images of dogs might be composed of very different sets of lines and curves, but both images will contain fur, a dog face, and a tail. Because there are more irrelevant features represented in lower layers, the percentage of polytope boundaries that relate features that are relevant to that class is smaller than between features represented in higher layers. 

\begin{figure}[!ht]
    \centering
    \includegraphics[width=0.6\linewidth]{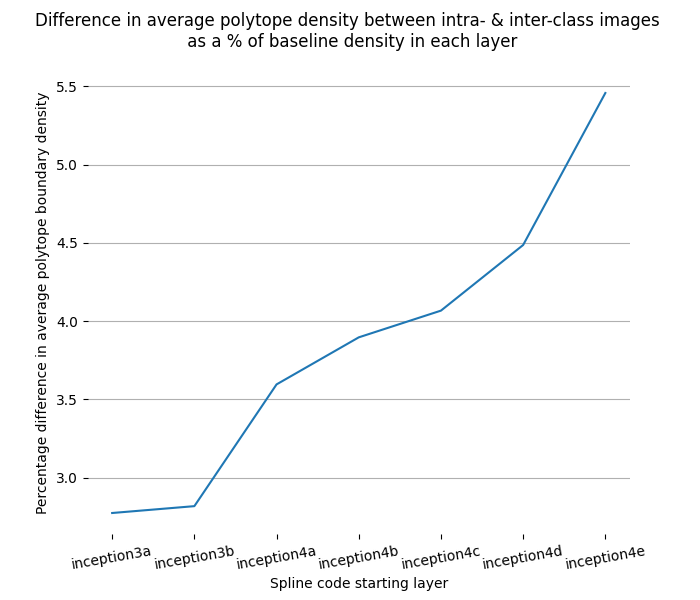}
    \caption{The difference between the normalized polytope density between intra- and inter-class images gets larger in layers closer to the output.}
    \label{fig:Fig19}
\end{figure}

\FloatBarrier

\subsubsection{Prediction 3: Polytopes define when feature-directions are on- and off-distribution}

One of the responses to the scaling activations experiments that we’ve encountered is that we’re being unfair to the networks: We shouldn’t expect their semantics to remain intact so far outside of their typical distribution. We agree! That there exists such a distribution of validity is, in fact, a central motivation for looking at networks through the polytope lens. 

The features-as-directions hypothesis doesn’t by itself make claims about the existence of a distribution of semantic validity because it assumes that representations are linear and therefore globally valid. The polytope lens predicts that scaling an activation vector will change the semantics of a given direction only when it crosses many polytope boundaries. It makes this prediction because the larger the distance between two polytopes, the more different (in expectation) is the transformation implemented by them. Polytopes boundaries thus suggest a way to identify the distribution of semantic validity. 

Is this the case empirically? Partially. When we plot the local polytope density in the region near the scaled vector, we see that there is a characteristic peak between the activation vector and the origin. This peak occurs even for activation directions defined by Gaussian noise, but is absent in untrained networks (Appendix B). There appears to be a ‘shell’ of densely packed polytope boundaries surrounding the origin in every direction we looked. We’re not completely sure why polytope boundaries tend to lie in a shell, though we suspect that it’s likely related to the fact that, in high dimensional spaces, most of the hypervolume of a hypersphere is close to the surface. Scaling up the activation, we see that the vector crosses a decreasing number of polytope boundaries. This is what you’d expect of polytope boundaries that lie near the origin and extend to infinity; as a result, polytopes further from the origin will be made from boundaries that become increasingly close to being parallel. Therefore a vector crosses fewer polytope boundaries as it scales away from the center. We nevertheless see plenty of class changes in regions that are distant from the origin that have low polytope density. This wasn’t exactly what the polytope lens predicted, which was that dense polytope boundaries would be located where there were class changes. Instead we observed dense polytope boundaries as we scale \emph{down} the activity vector and not as we scale it up. It appears that polytope boundaries only demarcate the inner bound of the distribution where a given direction means the same thing. That class changes can be observed for large magnitude activation vectors despite a low polytope boundary might simply reflect that it’s easier for large magnitude activations to move large distances when the transformations they undergo are small.

\begin{figure}[!ht]
    \centering
    \includegraphics[width=0.8\linewidth]{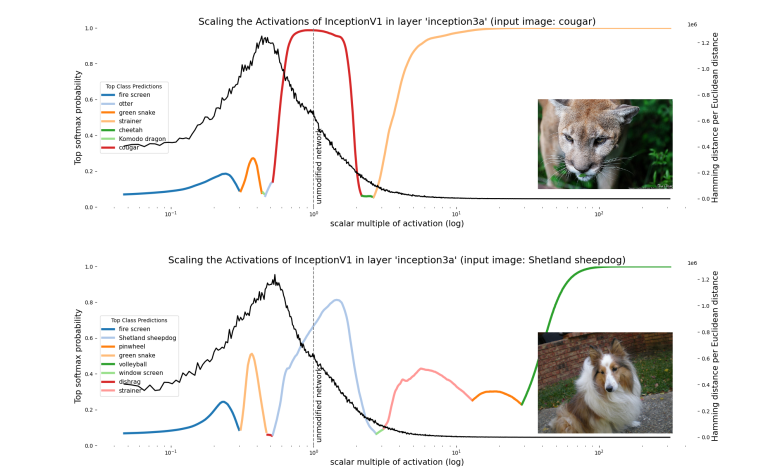}
    \caption{The polytope density (black line) overlaying the class logits (coloured lines) for two images where the activation in a hidden layer - inception3a - is scaled. Polytope density peaks around halfway between the unscaled activation vector and the origin.}
    \label{fig:Fig20}
\end{figure}

So polytope boundaries reflect - to some extent - the semantics learned by the network; they capture transformational invariances in the network, reflect feature boundaries, and seem to demarcate the inner bound of where feature-directions should be considered on- or off-distribution. They also seem to be involved in ``encoding'' features from raw data. Polytopes thus have many excellent properties for describing what is going on inside neural networks - but, as we will discuss in the next section, it's not clear how to harness polytopes to create \href{https://transformer-circuits.pub/drafts/toy_model_v2/index.html}{Decomposable} descriptions of the features in a network. Whilst studying neural networks through their polytope regions is a more 
``complete'' description in some sense, it does not (so far) let us understand network representations in terms of features that can be understood independently. 

\section{Discussion}

Our effort to account for nonlinearities in neural networks has forced us to consider not just the direction of neural activations, but also their scale. This is because nonlinearities behave differently at different activation scales. Polytopes offer a way to think about how networks use nonlinearities to implement different transformations at different activation scales. But with many neurons comes exponentially many polytopes. Spline codes present a scalable way to talk about the exponential number of polytopes in neural networks since we can talk about 
``groups'' or ``clusters'' of spline codes instead of individual codes.

Unfortunately, accounting for nonlinearities in this way has cost us rather a lot. Instead of dealing with globally valid feature directions, we now deal with only locally valid feature directions in activation space. By studying the structure of spline codes rather than the structure of activations, polytopes offer us the ability to identify regions of activation space that have roughly similar semantics. Are the costs worth the gains? 

The short answer is that we’re not sure. The polytope lens is a way to view neural networks that puts nonlinearities front and center; but if neural networks use primarily linear representations (as hypothesized by \cite{elhage2022superposition}), then such a nonlinearity-focused perspective could potentially offer relatively little compared to a purely linear perspective, since the abstraction of a globally valid feature direction will not be particularly leaky. The lesson we take from observations of superposition and polysemanticity is that networks are often not operating in the linear regime; they suggest that they are making nontrivial use of their nonlinearities to suppress interference from polysemantic directions. This is also suggested by the empirical performance of large networks which substantially exceeds the equivalent purely linear models. It therefore appears that we need a way to account for how different regions of activation space interact differently with nonlinearities and how this affects the semantics of the network’s representations.

We ultimately think that mechanistic descriptions of networks with superposition which take nonlinearity into account will look somewhat different from previous mechanistic descriptions that tended to assume linearity \citep{elhage2022superposition}. The polytope lens might represent an important component of such descriptions, but we’re in no way certain. If it were, what might mechanistic descriptions of neural networks look like through the polytope lens? 

\begin{figure}[!ht]
    \centering
    \includegraphics[width=0.8\linewidth]{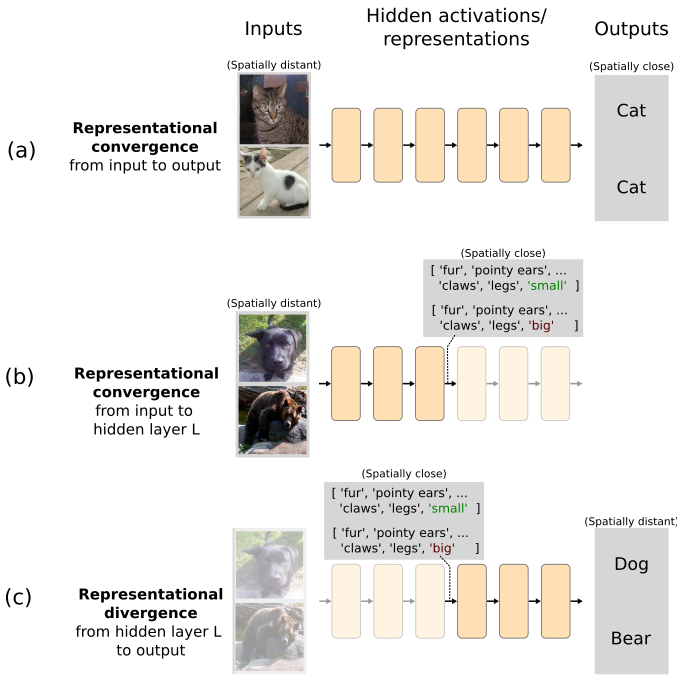}
    \caption{Representational flow between polytope regions might be a useful notion in mechanistic descriptions of neural networks.}
    \label{fig:Fig21}
\end{figure}

We think a potentially important idea for describing what neural networks have learned might be ‘\textbf{representational flow}’ between polytope regions. The input space of a layer may have regions that are semantically similar yet spatially distant and the job of the network is to learn how to project these spatially distant points to similar regions of output space. For example, the two images of cats in Figure~\ref{fig:Fig21}(a) below are distant in input space yet semantically similar in output space; the network performs \textbf{representational convergence} between representations in the input and output spaces. Representational convergence may also happen between arbitrary layers, such as between the input space and layer L if the inputs happen to share features that are represented in that layers’ semantic space (Figure~\ref{fig:Fig21}(b)). The converse is also possible: A network implements \textbf{representational divergence} if spatially similar inputs are semantically different from the perspective of the network at layer L (Figure~\ref{fig:Fig21}(c)). In order to implement representational \emph{convergence}, different polytopes need to have affine transformations that project them to similar parts of the space in later layers. Conversely, representational \emph{divergence} requires transformations that project nearby regions of activation space to distant regions in the spaces of later layers. Networks achieve both of these things by having the right affine transformations associated with the polytope regions involved. Nonlinearities mean that vectors that have an identical direction but different scales can take different pathways through the network. The benefit of thinking about networks in terms of representational flow is that it therefore allows us to talk about the effects of nonlinearities on activation directions of different scales. 

Recent work on superposition by \cite{elhage2022superposition} argues that models with superposition will only be understood if we can find a sparse overcomplete basis (or if we remove superposition altogether, an option we don’t consider here). Finding this basis seems like a crucial step toward understanding, but we don’t think it’s the full story. Even if we could describe a layer’s input features in terms of a sparse overcomplete basis, each combination of those sparse feature directions will have different patterns of interference which each interact differently with the nonlinearities. Thus, the elements of the sparse basis that are active will vary depending on the input vector; we therefore haven’t found a way around the issue that nonlinearities force us to use local, rather than global, bases.

Consequently, for most combinations it’s hard to predict exactly which activations will be above threshold without calculating the interference terms and observing empirically which are above or below threshold; this is a problem for mechanistic interpretability, where we’d like to be able to mentally model a network’s behavior without actually running it. Therefore, a sparse overcomplete basis by itself wouldn’t let us avoid accounting for nonlinearities in neural networks. Introducing assumptions about the input distribution such that interference terms are always negligibly small might however let us make predictions about a network’s behavior without adding schemes, like polytopes, that attempt to account for nonlinearities. 

Our work is more closely related to the search for an overcomplete basis than it might initially appear. Clustering activations can be thought of as finding a $k$-sparse set of features in the activations where $k=1$ (when $k$ is the number of active elements). In other words, finding $N$ clusters is equivalent to finding an overcomplete basis with $N$ basis directions, only one of which can be active at any one time. This clearly isn’t optimal for finding decomposable descriptions of neural networks; ideally we’d let more features be active at a time i.e. we’d like to let $k>1$, but with clustering $k=1$. But clustering isn’t completely senseless - If every combination of sparse overcomplete basis vectors interacts with nonlinearities in a different way, then every combination \emph{behaves} like a different feature. Fortunately, even if it were true that every combination of sparse overcomplete features interacted with nonlinearities in a different way, their interactions almost definitely have statistical and geometric structure, which we might be able to understand. Overcomplete basis features will be one component of that structure, but they don’t account for scale; polytopes do. A path toward understanding superposition in neural networks might be an approach that describes it in terms of an overcomplete basis \emph{and} in terms of polytopes. A potential future research direction might therefore be to find overcomplete bases in spline codes rather than simply clustering them. This might be one way to decompose the structure of representational flow into modules that account for both activation directions as well as activation scale.

Many other questions remain unaddressed in this post. We think they will be important to answer before the polytope lens can be used in as many circumstances as the features-as-directions perspective has been.

\begin{itemize}
    \item \textbf{Fuzzy polytope boundaries with other activations} - For the sake of simplicity, we’ve been assuming that the networks discussed in the article so far have used piecewise linear activation functions such as ReLU. But many networks today, including large language models, often use smooth activations such as GELU and softmax, which mean that their polytopes won’t really be polytopes - their edges will be curvy or even ‘blurred’. Some prior work exists that extends the polytope lens to such activations \citep{balestriero2018}. See Appendix C for further discussion.
    \item \textbf{How do we extend the polytope lens to transformers?} Specifically, how should we talk about polytopes when attention between embedding vectors makes activations (and hence polytopes) interact multiplicatively across sequence positions? 
    \item \textbf{How do adversarial examples fit into this picture?} Are adversarial examples adversarial because they perturb the input such that it crosses many polytope boundaries (polytope ridges)? And can we use this potential insight in order to make networks less susceptible to such attacks? 
\end{itemize}

\section{Related work}

\subsection{Interpreting polytopes, single neurons, or directions}

The geometric interpretation of ReLU networks was, to our knowledge, first laid out by \cite{nair2010}, who note that each unit corresponds to a hyperplane through the input space, and that $N$ units in concert can create $2^N$ regions (what we call polytopes), each of which can be viewed as a separate linear model. \cite{pascanu2014} undertook a more detailed theoretical analysis of the number of these linear regions in ReLU models.

The fact that each of these regions could be identified as a unique \emph{code}, which can then be used for interpretability analysis and clustering, was explored by \cite{srivastava2015}, who studied a small MNIST network by clustering the codes at its final layer. 

That these regions take the form of convex polytopes is also not a novel concept, and has been explored in a number of prior works \citep{balestriero2018a,novak2018,hanin2019b,rolnick20a,xu2022traversing}. In this writeup, we have relied particularly heavily on conceptualizing DNNs as compositions of \emph{max-affine spline operators}, as introduced in \cite{balestriero2018a}, and expanded upon in a series of further works \citep{balestriero2018,balestriero2019}. 

However, in much of the wider interpretability field – particularly in papers focused on interpretability in language models – this point of view has gone largely unnoticed, and interpretation efforts have tended to try to identify the role of single neurons or linear combinations of neurons (directions). Interpretable neurons have been noted fairly widely in various works focusing on vision models \citep{szegedy2014,Bau2017}. Interpretable directions were also a central focus of the \href{https://distill.pub/2020/circuits/}{Circuits Thread}, \citep{cammarata2020thread}, where they used knowledge built up from interpreting neurons in early layers of inceptionv1 to hand code curve detectors that, when substituted for the curve detectors in the original network, induced minimal performance loss.  

Interpretable single neurons have also been found in language models \citep{geva2021,durrani2020,dai2022,elhage2022superposition}, although monosemantic neurons seem comparatively less common in this class of model. \href{https://arxiv.org/abs/2104.07143}{\emph{An Interpretability Illusion for BERT}} \citep{bolukbasi2021}, highlighted the fact that the patterns one might see when inspecting the top-$k$ activations of some neuron may cause us to spuriously interpret it as encoding a single, simple concept, when in fact it is encoding for something far more complex. They also noted that many directions in activation space that were thought to be globally interpretable may only be locally valid.

\subsection{Polysemanticity and Superposition}

The earliest mention of polysemanticity we could find in machine learning literature was from \cite{nguyen2016}. In their paper they identify the concept of \emph{multifaceted} neurons. That is, neurons which fire in response to many different types of features. In this work, we define \emph{polysemantic} neurons as neurons which fire in response to many different \emph{unrelated} features, and they identify an example of this in their supplementary material (Figure S5).

Work by \cite{olah2017feature}, \href{https://distill.pub/2017/feature-visualization/}{Feature Visualization}, identified another way to elicit polysemantic interpretations and helped to popularize the idea. They note that, as well as there being neurons which represent a single coherent concept, \emph{“\dots there are also neurons that represent strange mixtures of ideas. Below, a neuron responds to two types of animal faces, and also to car bodies. Examples like these suggest that neurons are not necessarily the right semantic units for understanding neural nets.”}

\begin{figure}[!ht]
    \centering
    \includegraphics[width=0.85\linewidth]{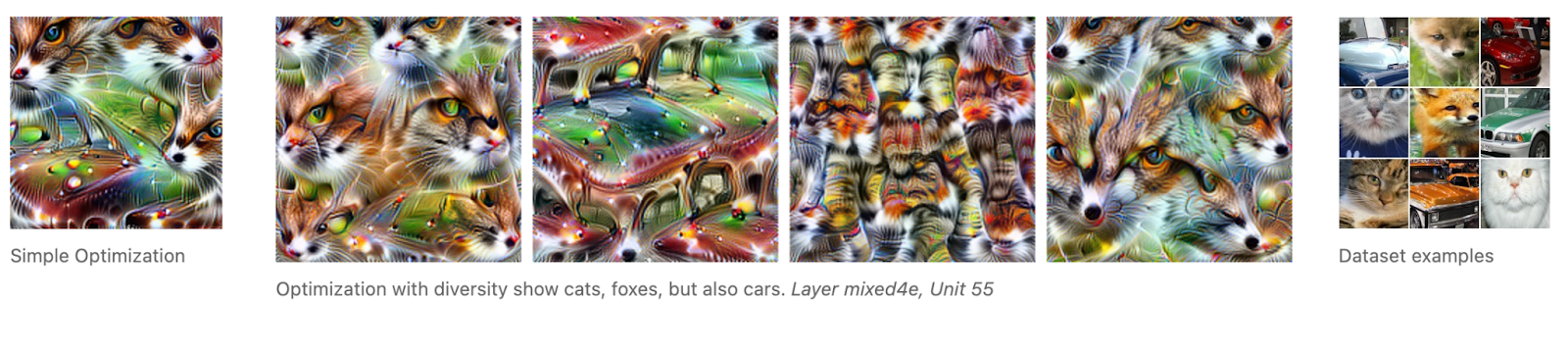}
    \caption{Image from \cite{olah2017feature} depicting a polysemantic neuron.}
    \label{fig:Fig22}
\end{figure}

Even before this, the possibility that individual neurons could respond to multiple features was discussed in some early connectionist literature, including \cite{hinton1981}. In neuroscience, polysemanticity is usually called ‘mixed selectivity’. Neuroscience has only in the last decade or two developed the tools required to identify and study mixed selectivity. Since then, it has been the subject of increasing attention, especially its role in motor- and decision- neuroscience \citep{churchland2007,rigotti2013,mante2013}. For a review of mixed selectivity in neuroscience, see \cite{fusi2016}.  

Recent work from \cite{elhage2022superposition} sheds light on a phenomenon that they term “superposition”. Superposition occurs when a neural network represents more features than it has dimensions, and the mapping from features to orthogonal basis directions can no longer be bijective. This phenomenon is related to, but not the same as polysemanticity; it may be a cause of some of the polysemantic neurons we see in practice. They investigate toy models with non-linearities placed at the output layer, and show that superposition is a real phenomenon that can cause both mono- and polysemantic neurons to form. They also describe a simple example of computation being performed on features in superposition. Finally, they reveal that superposition can cause a different type of polytope to form - in their toy model, features are organized into geometric structures that appear to be a result of a repulsive force between feature directions which acts to reduce interference between features. It’s worth emphasizing that the polytopes discussed in their work aren’t the same kind as in ours: For one, our polytopes lie in activation space whereas theirs lie in the model weights. Perhaps a more fundamental divergence between Elhage et al.’s model and ours is the assumption of linearity - the idea that features are represented by a single direction in activation space. As we explained in earlier sections, we believe that assuming linearity will yield only partial mechanistic understanding of nonlinear networks. While globally valid feature directions would simplify analysis, in practice we struggle to see a way around nonlinearity  by assuming linear representations. 

\begin{ack}
This work benefited from feedback from many staff at Conjecture including Adam Shimi, Nicholas Kees Dupuis, Dan Clothiaux, Kyle McDonell. Additionally, the post also benefited from inputs from Jessica Cooper, Eliezer Yudkowsky, Neel Nanda, Andrei Alexandru, Ethan Perez, Jan Hendrik Kirchner, Chris Olah, Nelson Elhage, David Lindner, Evan R Murphy, Tom McGrath, Martin Wattenberg, Johannes Treutlein, Spencer Becker-Kahn, Leo Gao, John Wentworth, and Paul Christiano and from discussions with many other colleagues working on interpretability. 
\end{ack}

\small
\bibliography{references}

\clearpage
\appendix

\section{Polytope density while interpolating between activations caused by images}

\begin{figure}[!ht]
    \centering
    \includegraphics[width=0.85\linewidth]{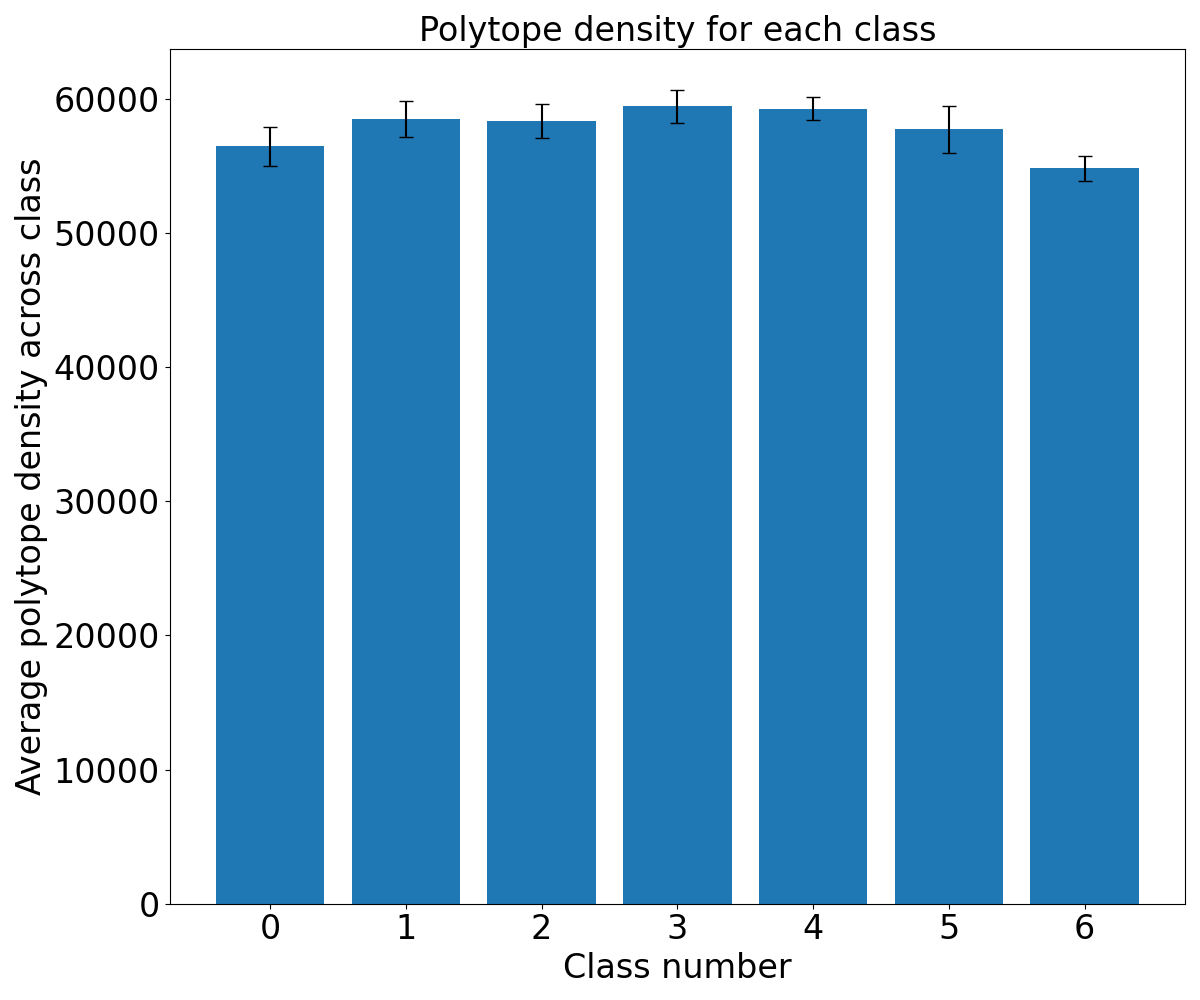}
    \caption{The polytope density for each class during a spherical interpolation between the activations caused by images of different classes in inception3a. The polytope codes were computed from the activations at layer 3a to the output layer. The polytope density was estimated by sampling 150 random points around each position during the interpolation and computing the number of polytopes passed through versus the euclidean distance. The interpolation path passes through multiple other classes. We see that the polytope density is highest in the intermediate region where there is much class change between intermediate classes. This trend is relatively weak, however. This provides tentative evidence in favor of the relationship between polytope density and semantic change in representation.}
    \label{fig:Fig23}
\end{figure}

\begin{figure}[!ht]
    \centering
    \includegraphics[width=0.85\linewidth]{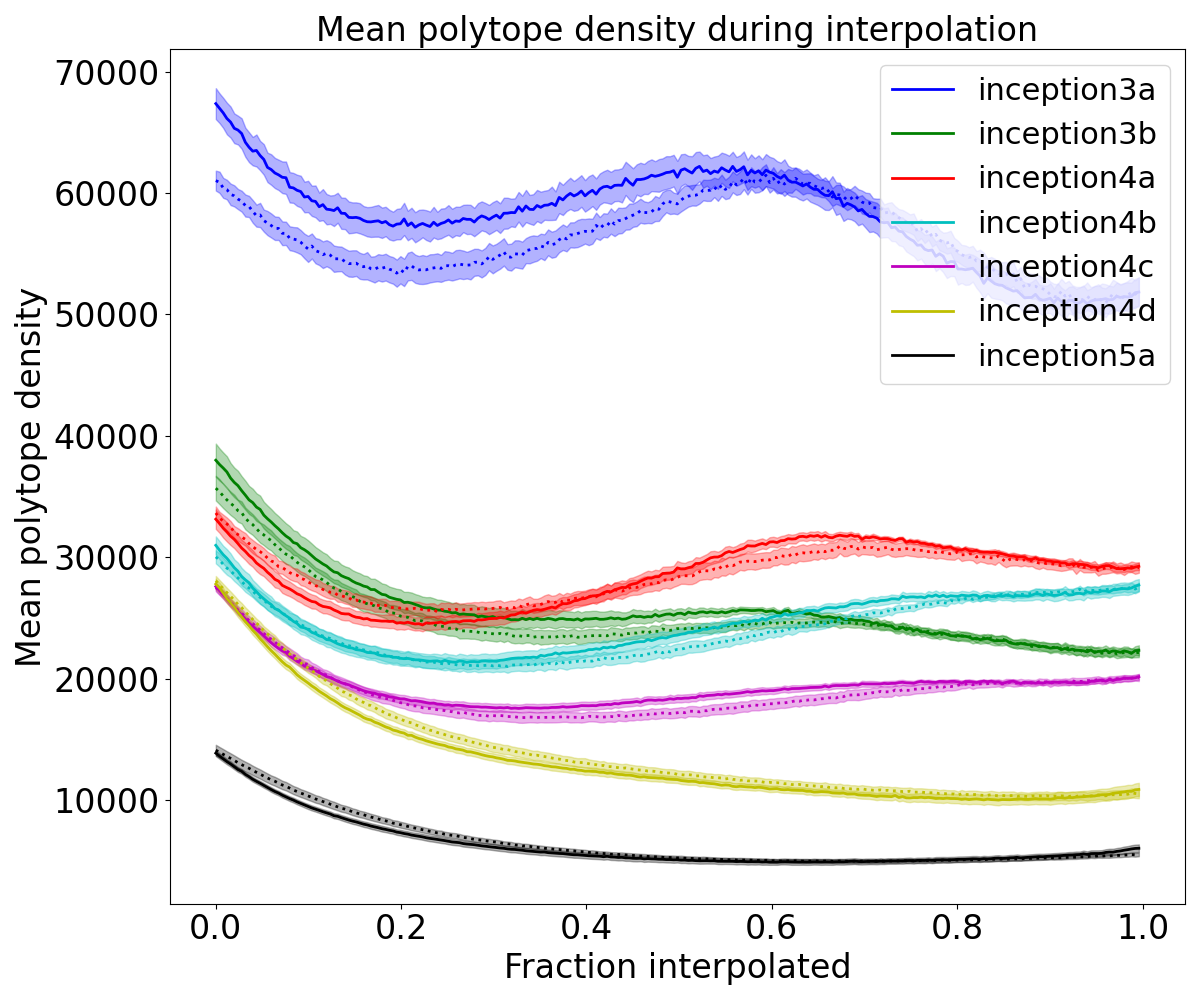}
    \caption{Mean polytope density (averaged over 200 image interpolations) by spherically interpolating between different class examples on each layer. Dotted lines represent the mean interpolation between images of the same class and solid lines represent the mean interpolation between images of different classes. The shaded regions represent the standard error of the mean. The polytope codes were computed from the embedding at the labeled layer to the output space. For lower imagenet layers, where the class labels are less informative of the semantic features, we see that the polytope density exhibits a double dip phenomenon where it increases when about half interpolated, indicating that interpolation leaves the manifold of typical activations. For later layers, the polytope density decreases and the curve flattens during interpolation implying that at these layers there are more monosemantic polytopes and the class labels are more representative of the feature distribution.}
    \label{fig:Fig24}
\end{figure}

\clearpage

\section{Scaling activation vectors and plotting polytope density}

\subsection*{Untrained network}
\begin{figure}[!ht]
    \centering
    \includegraphics[width=0.9\linewidth]{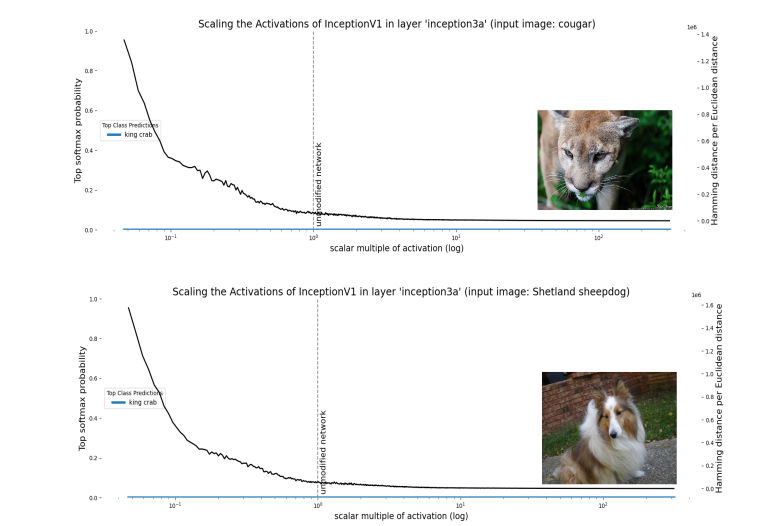}
    \caption{Note that the unchanging class in the untrained network is due to a phenomenon that resembles ‘rank collapse’: Even though the input and early activations are different, the activations of the untrained network converge on the same output. We believe this might due to a quirk for our variant of InceptionV1 (perhaps its batchnorm), but we haven’t investigated why exactly this happens.}
    \label{fig:Fig25}
\end{figure}

\subsection*{With Gaussian noise activations}
\begin{figure}[!ht]
    \centering
    \includegraphics[width=0.9\linewidth]{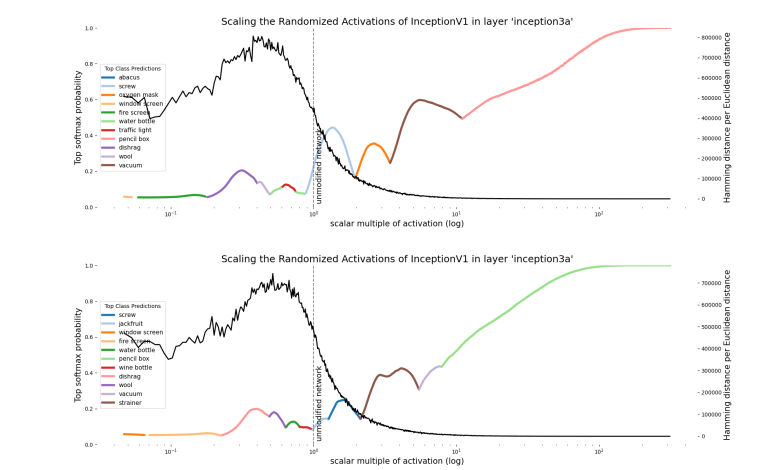}
    \caption{}
    \label{fig:Fig26}
\end{figure}

\clearpage
\section{Mathematical account of neural networks as max affine spline operators (MASOs)}

In the below section we give an account of some recent theory from \cite{Balestriero2018b} that links deep neural networks to approximation theory via spline functions and operators. More specifically, the authors describe deep neural networks with piecewise linear activation functions (like ReLU) as compositions of \emph{max-affine spline operators (MASOs)}, where each layer represents a single MASO. A MASO is an \emph{operator} composed of a set of individual \emph{max-affine spline} functions (MASs), one for each neuron in a given nonlinear layer. 

We won’t go too deep into spline approximation theory here, but you can think of a spline function approximation in general as consisting of a set of partitions $\Omega^R$ of the input space, with a simple local mapping in each region. The \emph{affine} part means that this mapping consists of an affine transformation of the input in a given region: 
\[
a_rx + b_r \text{ for } r=1,\dots, R
\]
The \emph{max} part means that, instead of needing to specify the \emph{partition region} of our input variable in order to determine the output, we can simply take the maximum value when we apply the entire set of affine transformations for each region:
\[
z(x) = \displaystyle\max_{r=1,\dots,R} a_r x + b_r
\]
A visual example is helpful to understand why this works. Suppose we have a spline approximation function with $R=4$ regions:
\begin{center}
    \includegraphics[width=0.9\linewidth]{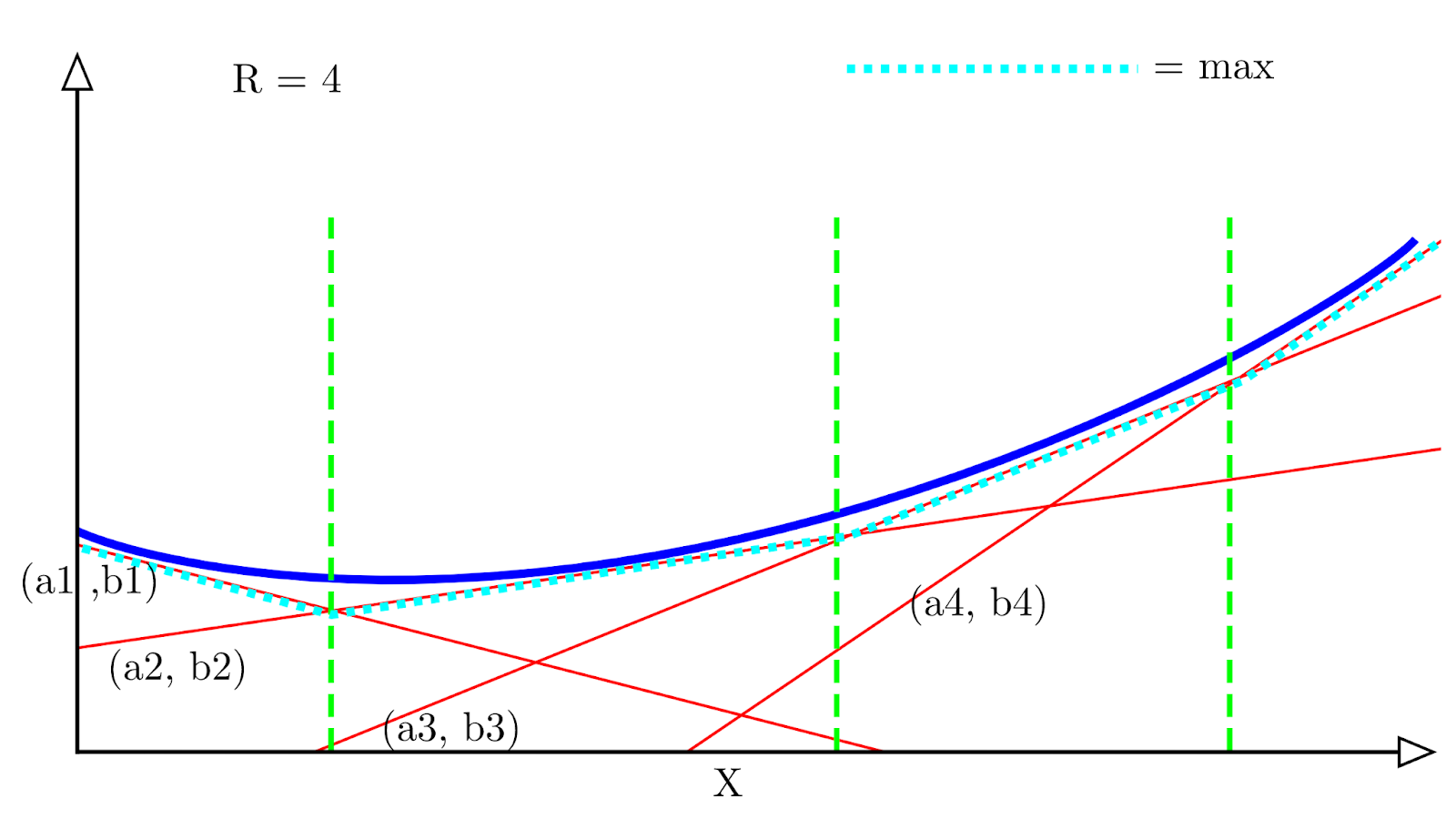}
\end{center}

Each red line represents a single spline with a corresponding affine transformation $(a_r,b_r)$, and the dotted light blue line represents the maximum value of all the affine transformations at each $x$ location. We can see that it follows an approximation of the convex function (in dark blue).

A single ReLU unit can be expressed as a special case of \emph{max-affine spline} with $R=2$ regions:
\[
\operatorname{relu}(x) = \max_{r=1,2} a_r x + b_r
\]
Where $(a_1,b_1)=(0,0)$ and $(a_2,b_2)=(W_i,b_i)$, which are the weight and bias vectors for a given neuron. An entire ReLU layer can then be seen simply as a concatenation of $d$ of these $R=2$ MASs, where $d$ is the width of the layer – this is our MASO.

This becomes slightly more complicated for smooth activation functions like GELU and Swish. But, fortunately, in a \href{https://arxiv.org/pdf/1810.09274.pdf}{later paper} the same authors extend their framework to just such functions. In summary - smooth activation functions must be represented with a \emph{probabilistic} spline code rather than a one-hot binary code. The corresponding affine transformation at the input point is then a linear interpolation of the entire set of affine transformations, weighted by the input point’s probability of belonging to each region.

\clearpage
\section{Note on Terminology of Superposition, Interference, and Aliasing}

The concepts referred to by the terms ‘superposition’ and ‘interference’ \cite{elhage2022superposition} have parallel names in other literature. We provide this footnote with the hope of inspiring links between mechanistic interpretability and related results in signal processing, systems theory, approximation theory, physics, and other fields. 

The \href{https://en.wikipedia.org/wiki/Superposition_principle}{superposition principle} in the theory of linear systems refers to the fact that states of or solutions to a linear system may be added together to yield another state or solution. For example, solutions to linear wave equations may be summed to yield another solution. In this sense, superposition tells us that we can mathematically deduce the action of a system on any input from its action on a set of orthogonal basis vectors. This usage clashes with its usage in the mechanistic interpretability literature so far, where it has often been used to refer to systems without such a decomposition property. ‘Interference’ generally refers to superposition applied to linear waves. Specifically, the components of two waves interfere with each other, but orthogonal components within a wave do not.

The notion of ‘superposition’ and ‘interference’ as used in \cite{elhage2022superposition}, where different features fail to be completely independent and inhibit correct measurements is similar to the idea of \href{https://en.wikipedia.org/wiki/Aliasing}{aliasing} in other literatures. The term 'aliasing' originates in signal processing. In that context, aliasing arose from the indistinguishability of waves of different frequencies under discrete sampling schemes. Aliasing has come to refer more generally to the phenomenon in which a set of desired quantities (e.g. features) fails to be orthogonal with respect to a measurement basis. If we wish to determine the value of $n$ features from $k << n$ measurements, some sets of feature values may yield the same measurements. In the case of sampling waves, high-frequency waves may appear the same as low-frequency waves. In the case of approximating functions from k many sample points, high-degree polynomials may take the same values on those k points \citep[see][Chapter 4 for a discussion in the case of Chebyshev interpolation]{trefethen2019}. In image processing, \href{https://en.wikipedia.org/wiki/Spatial_anti-aliasing}{anti-aliasing} is used to deal with visual artifacts that come from high-frequency components being indistinguishable from lower frequency components. 

Quantum mechanics uses the conventions we have described. A quantum system with two possible classical states $|0>$ and $|1>$ has its quantum state described as an orthogonal superposition of the form $a|0>+b|1>$ where $a$ and $b$ are complex numbers. The two classical states do not ‘interfere’ with each other. Rather, two independent quantum systems may additively interfere with corresponding orthogonal components interfering. Interference and superposition in this context are not referring to entanglement. Just as we may represent $(|0>+|1>)/\sqrt{2}$ as a superposition of the states $|0>$ and $|1>$, we may also represent the state $|0>$ as a superposition of the states $(|0>+|1>)/\sqrt{2}$ and $(|0>-|1>)/\sqrt{2}$. The important detail regarding ‘superposition’ is the additivity, not the particular choice of classical states for our representation. The \href{https://en.wikipedia.org/wiki/Quantum_superposition}{quantum harmonic oscillator} has eigenstates (orthogonal basis vectors for the system) described by \href{https://en.wikipedia.org/wiki/Hermite_polynomials}{Hermite polynomials}. If we approximate the Hermite polynomials with an asymptotic approximation, we will observe aliasing due to the failure of our approximation to be perfectly orthogonal.

\clearpage
\section{Examples of Text Clusters from GPT2-Small}

\subsection*{Spline code clusters (computed with codes from layer L -> output):}

\begin{figure}[!ht]
    \centering
    \includegraphics[width=0.85\linewidth]{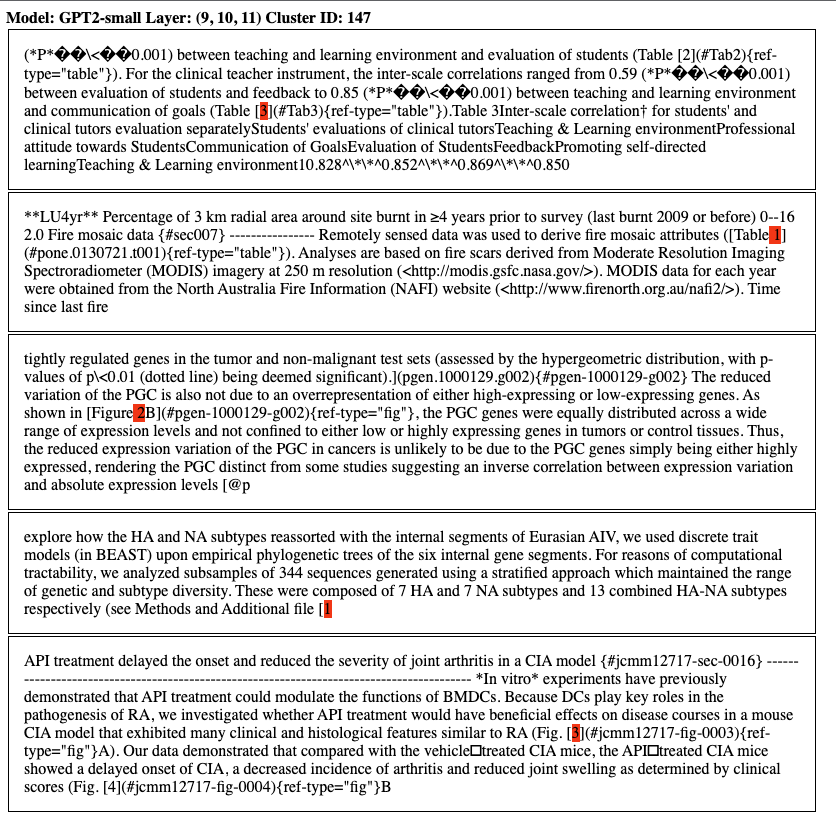}
    \caption{A cluster responding to figure and table references in latex documents.}
    \label{fig:Fig28}
\end{figure}

\begin{figure}[!ht]
    \centering
    \includegraphics[width=0.85\linewidth]{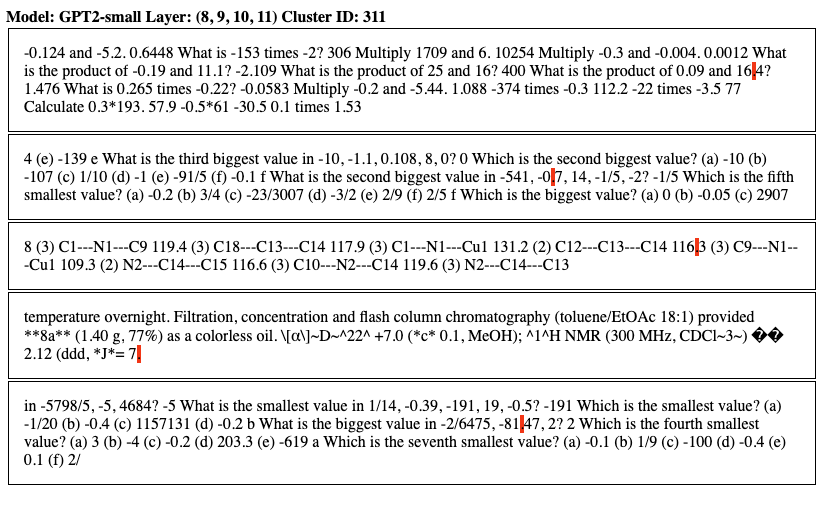}
    \caption{A cluster responding to decimal points in numbers.}
    \label{fig:Fig29}
\end{figure}

\begin{figure}[!ht]
    \centering
    \includegraphics[width=0.85\linewidth]{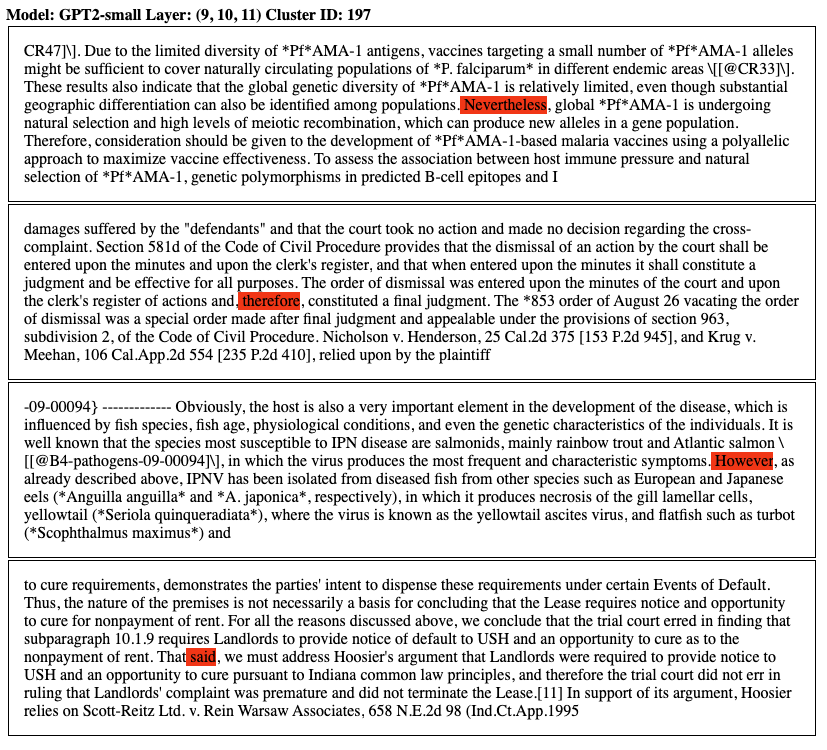}
    \caption{A cluster responding to words followed by commas (or conjunctive pronouns?).}
    \label{fig:Fig30}
\end{figure}

\begin{figure}[!ht]
    \centering
    \includegraphics[width=0.85\linewidth]{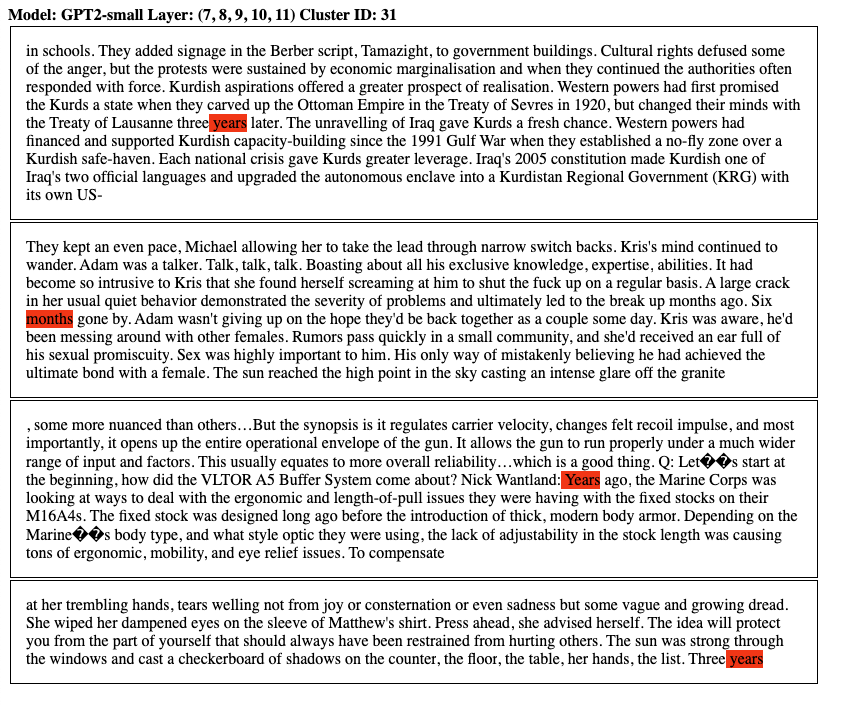}
    \caption{A cluster responding to spans of time.}
    \label{fig:Fig31}
\end{figure}

\clearpage

\subsection*{Activation clusters:}

\begin{figure}[!ht]
    \centering
    \includegraphics[width=0.78\linewidth]{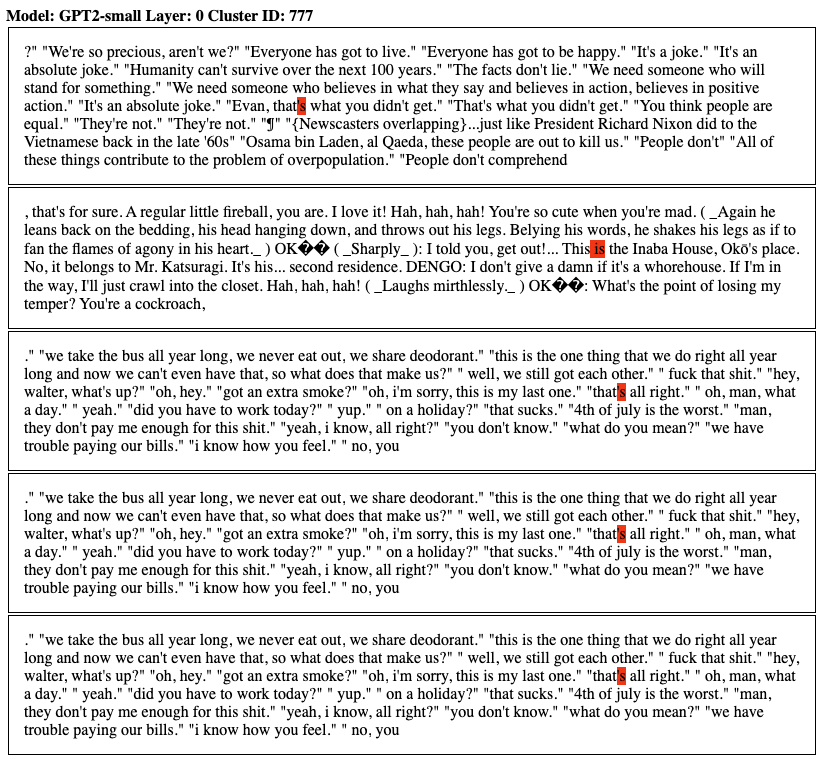}
    \caption{A 'detokenization' cluster that responds both to the word “is” and its contraction.}
    \label{fig:Fig32}
\end{figure}

\begin{figure}[!ht]
    \centering
    \includegraphics[width=0.78\linewidth]{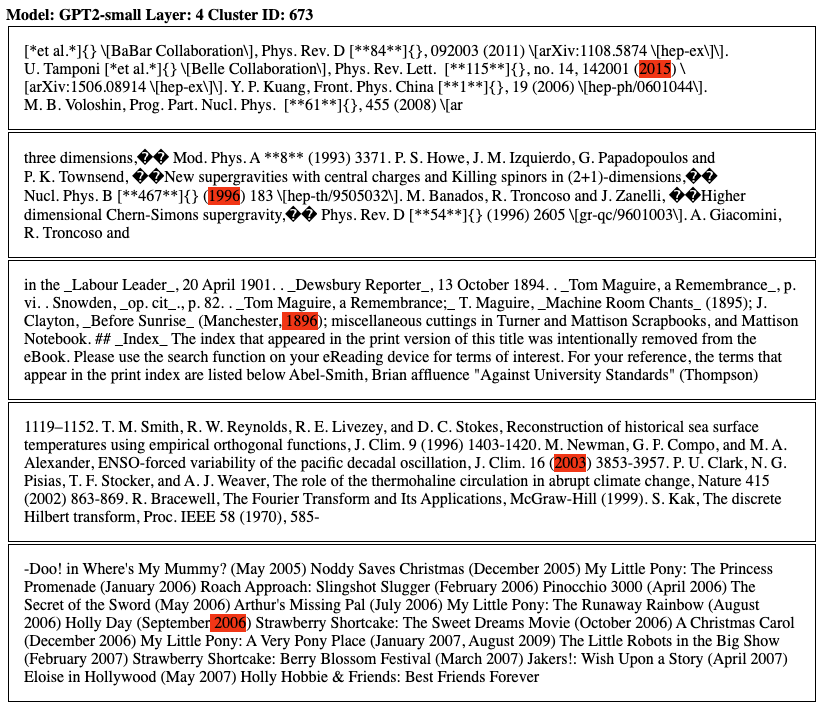}
    \caption{A cluster responding to dates.}
    \label{fig:Fig33}
\end{figure}

\begin{figure}[!ht]
    \centering
    \includegraphics[width=0.78\linewidth]{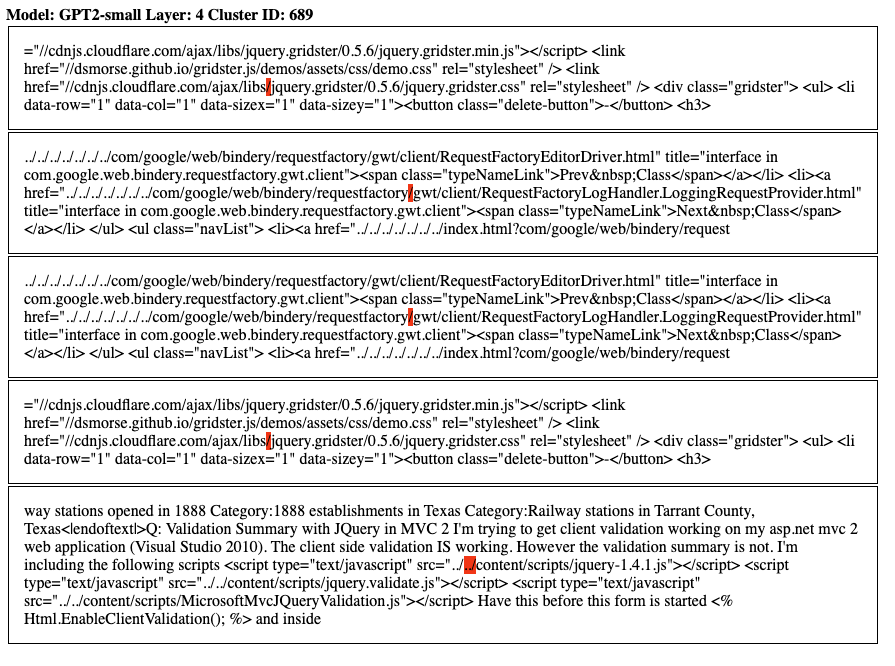}
    \caption{A cluster responding to forward slashes in file paths.}
    \label{fig:Fig34}
\end{figure}

\begin{figure}[!ht]
    \centering
    \includegraphics[width=0.78\linewidth]{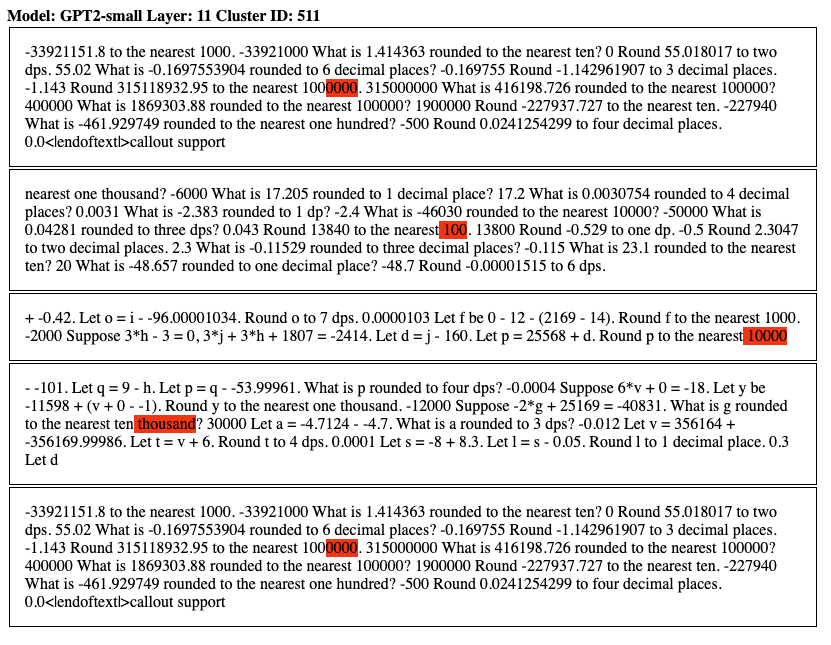}
    \caption{A cluster responding to multiples of ten (verbal and numeric).}
    \label{fig:Fig35}
\end{figure}

\end{document}